\documentclass[10pt,twocolumn,letterpaper]{article}

\usepackage{cvpr}
\usepackage{times}
\usepackage{epsfig}
\usepackage{graphicx}
\usepackage{amsmath}
\usepackage{amssymb}

\usepackage{wrapfig}
\usepackage{pdfpages}
\usepackage{sidecap}
\usepackage{colortbl}
\usepackage{overpic}
\usepackage{booktabs}
\usepackage{colortbl}
\usepackage[toc,page]{appendix}
\usepackage{placeins}
\usepackage{float}

\usepackage{rotating}
\usepackage[export]{adjustbox}
\usepackage[table]{xcolor}

\definecolor{red}{rgb}{0.95,0.4,0.4}
\definecolor{blue}{rgb}{0.4,0.4,0.95}
\definecolor{darkblue}{rgb}{0,0,0.8}
\definecolor{darkred}{rgb}{0.8,0,0}
\definecolor{darkgreen}{rgb}{0,0.5,0}
\definecolor{grey}{rgb}{0.6,0.6,0.6}
\definecolor{col1}{RGB}{232, 161, 148}
\definecolor{col2}{RGB}{148, 187, 232}

\newcommand{\ncite}[1]{{\color{red} [cite] }}

\usepackage{xspace}
\newcommand*{\ea}{et al.\@\xspace}

\usepackage{pifont}%
\newcommand{\xmark}{\ding{55}}%

\usepackage[pagebackref=true,breaklinks=true,letterpaper=true,colorlinks,bookmarks=false]{hyperref}

\cvprfinalcopy 

\def\cvprPaperID{9078} 

\ifcvprfinal\fi
\begin{document}

\title{Self-supervised Monocular Trained Depth Estimation using \\ Self-attention and Discrete Disparity Volume}

\author{Adrian Johnston \quad Gustavo Carneiro \\
	Australian Institute for Machine Learning \\
	School of Computer Science, University of Adelaide \\
	{\tt\small \{adrian.johnston, gustavo.carneiro\}@adelaide.edu.au}
}

\maketitle

\begin{abstract}
Monocular depth estimation has become one of the most studied applications in computer vision, where the most accurate approaches are based on fully supervised learning models. However, the acquisition of accurate and large ground truth data sets to model these fully supervised methods is a major challenge for the further development of the area. Self-supervised methods trained with monocular videos constitute one the most promising approaches to mitigate the challenge mentioned above due to the wide-spread availability of training data.  Consequently, they have been intensively studied, where the main ideas explored consist of different types of model architectures, loss functions, and occlusion masks to address non-rigid motion. In this paper, we propose two new ideas to improve self-supervised monocular trained depth estimation: 1) self-attention, and 2) discrete disparity prediction. Compared with the usual localised convolution operation, self-attention can explore a more general contextual information that allows the inference of similar disparity values at non-contiguous regions of the image.  Discrete disparity prediction has been shown by fully supervised methods to provide a more robust and sharper depth estimation than the more common continuous disparity prediction, besides enabling the estimation of depth uncertainty.  We show that the extension of the state-of-the-art self-supervised monocular trained depth estimator Monodepth2 with these two ideas allows us to design a model that produces the best results in the field in KITTI 2015 and Make3D, closing the gap with respect self-supervised stereo training and fully supervised approaches.

\end{abstract}




\begin{figure}
	\centering
	\includegraphics[clip,width=1.0\linewidth]{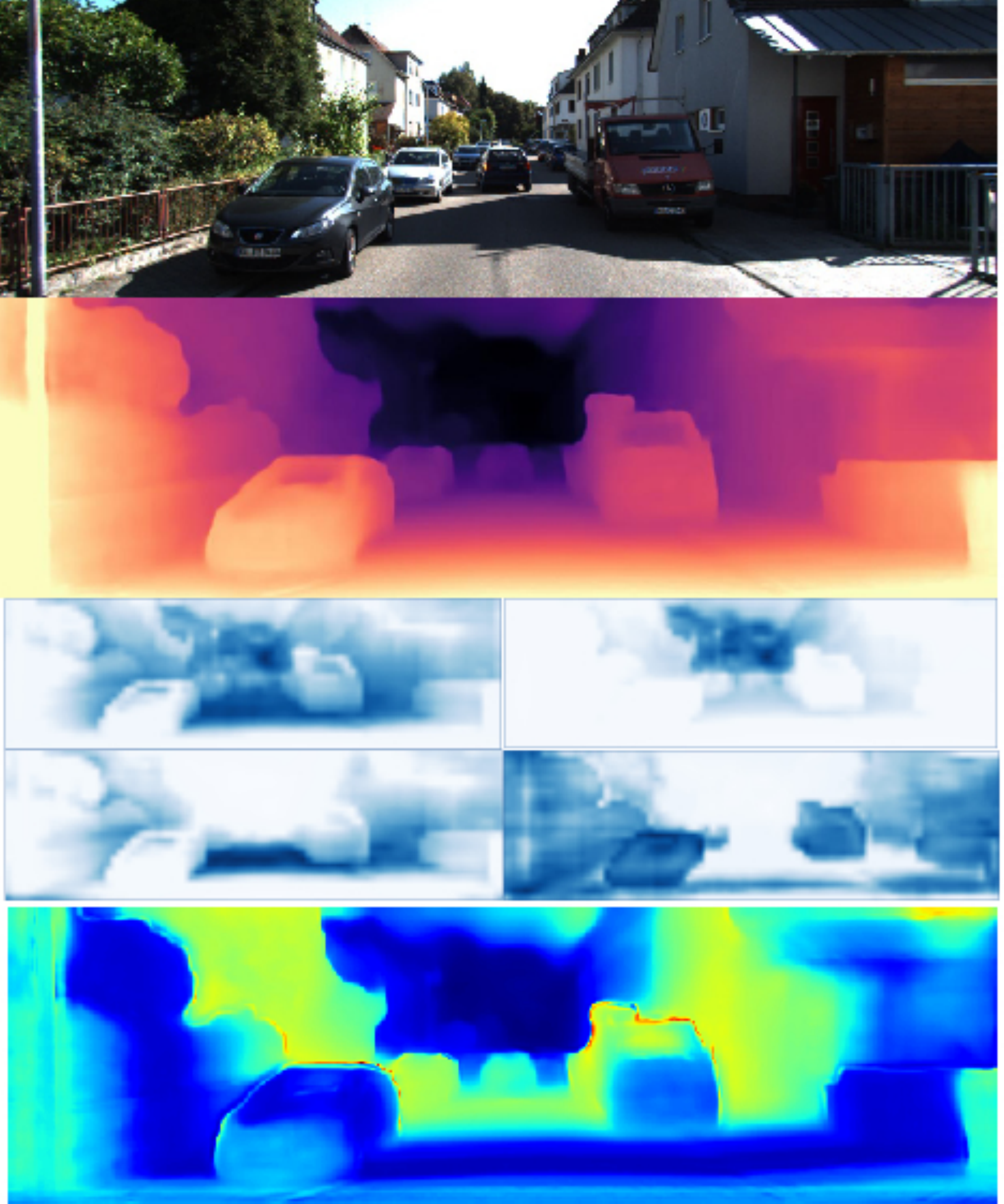}
	\caption{{\bf Self-supervised Monocular Trained Depth Estimation using Self-attention and Discrete Disparity Volume}. Our self-supervised monocular trained model uses self-attention to improve contextual reasoning and discrete disparity estimation to produce accurate and sharp depth predictions and depth uncertainties. 
    \emph{Top: input image}; 
    \emph{Middle Top: estimated disparity}; 
    \emph{Middle Bottom: samples of the attention maps produced by our system 
    (blue indicates common attention regions)}; 
    \emph{Bottom: pixel-wise depth uncertainty (blue: low uncertainty; green/red: high/highest uncertainty).}
    }
	\label{fig:front_figure}
\end{figure}

\vspace{-1pt}
\section{Introduction}
\label{sec:introduction}

\begin{figure*}
	\includegraphics[clip,width=1.0\linewidth]{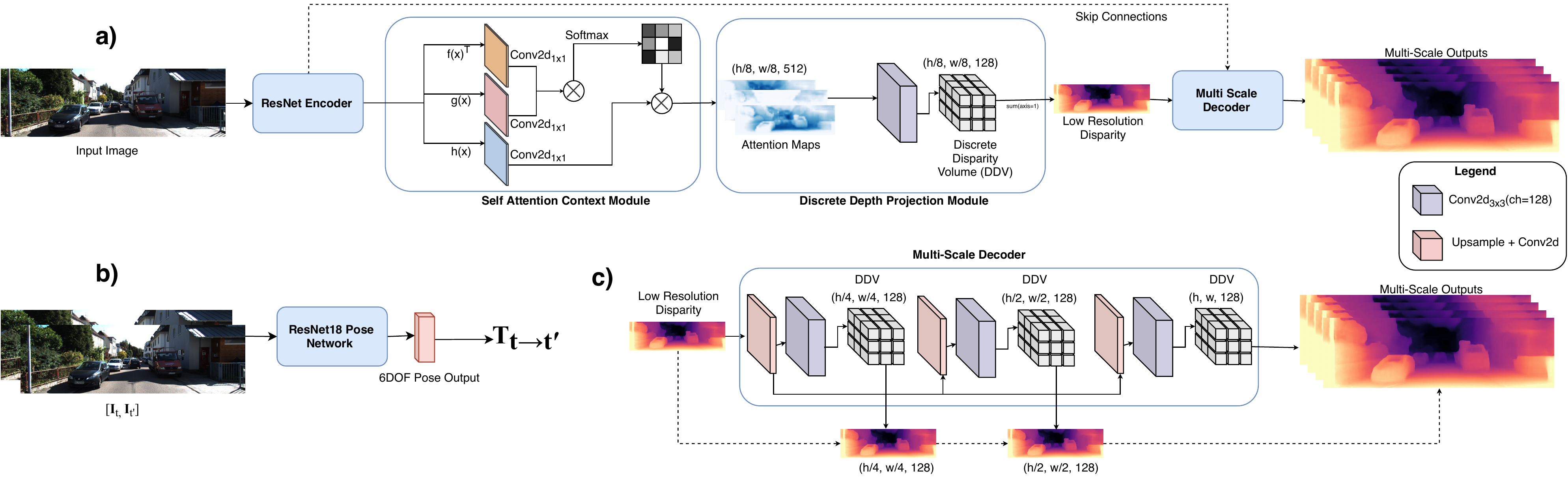}
	\caption{{\bf Overall Architecture} The image encoding processes is highlighted in part \emph{a)}.  
    The input monocular image is encoded using a ResNet encoder and then passed through the Self-Attention Context Module. The computed attention maps are then convolved with a 2D convolution with the number of output channels equal to the number dimensions for the Discrete Disparity Volume (DDV). The DDV is then projected into a 2D depth map by performing a \emph{softargmax} across the disparity dimension resulting in the lowest resolution disparity estimation (Eq.~\ref{eq:DDV_low_res}). In part \emph{b)} the
    pose estimator is shown, and part \emph{c)} shows more details of the Multi-Scale decoder. The low resolution disparity map is passed through successive blocks of UpConv (nearest upsample + convolution). The DDV projection is performed at each scale, in the same way as in the initial encoding stage. Finally, each of the outputs are upsampled to input resolution to compute the photometric reprojection loss. }
	\label{fig:arch}
\end{figure*}

Perception of the 3D world is one of the main tasks in computer/robotic vision. Accurate perception, localisation, mapping and planning capabilities are predicated on having access to correct depth information. Range finding sensors such as LiDAR or stereo/multi-camera rigs are often deployed to estimate depth for use in robotics and autonomous systems, due to their accuracy and robustness.  However, in many cases it might be unfeasible to have, or rely solely on such expensive or complex sensors. This has led to the development of learning-based methods~\cite{saxena2006learning, saxena2009make3d, karsch2014depth}, where the most successful approaches rely on fully supervised convolutional neural networks (CNNs)~\cite{eigen2014depth,eigen2015predicting, fu2018deep, guo2018learning, mayer2018makes}. While supervised learning methods have produced outstanding monocular depth estimation results, ground truth RGB-D data is still limited in variety and abundance when compared with the RGB image and video data sets available in the field. Furthermore, collecting accurate and large ground truth data sets is a difficult task due to sensor noise and limited operating capabilities (due to weather conditions, lighting, etc.). 

Recent studies have shown that it is instead possible to train a depth estimator in a self-supervised manner using synchronised stereo image pairs~\cite{garg2016unsupervised, godard2017unsupervised} or monocular video~\cite{zhou2017unsupervised}.
While monocular video offers an attractive alternative to stereo based learning due to wide-spread availability of training sequences, it poses many challenges. Unlike stereo based methods, which have a known camera pose that can be computed offline, self-supervised monocular trained depth estimators need to jointly estimate depth and ego-motion to minimise the photometric reprojection loss function~\cite{garg2016unsupervised, godard2017unsupervised}. Any noise introduced by the pose estimator model can degrade the performance of a model trained on monocular sequences, resulting in large depth estimation errors. Furthermore, self-supervised monocular training makes the assumption of a moving camera in a static (i.e., rigid) scene, which causes monocular models to estimate 'holes' for pixels associated with moving visual objects, such as cars and people (i.e., non-rigid motion). To deal with these issues, many works focus on the development of new specialised architectures~\cite{zhou2017unsupervised}, masking strategies~\cite{zhou2017unsupervised, monodepth2, vijayanarasimhan2017sfm, luo2018every}, and loss functions~\cite{godard2017unsupervised, monodepth2}. Even with all of these developments, self-supervised monocular trained depth estimators are less accurate than their stereo trained counterparts and significantly less accurate than fully supervised methods. 

In this paper, we propose two new ideas to improve self-supervised monocular trained depth estimation: 1) self-attention~\cite{non-local,vaswani2017attention}, and 2) discrete disparity volume~\cite{kendall2017end}.  
Our proposed self-attention module explores non-contiguous (i.e., global) image regions as a context for estimating similar depth at those regions.  Such approach contrasts with the currently used local 2D and 3D convolutions that are unable to explore such global context. The proposed discrete disparity volume enables the estimation of more robust and sharper depth estimates, as previously demonstrated by fully supervised depth estimation approaches~\cite{kendall2017end,liu2019neural}. Sharper depth estimates are important to improving accuracy, and increased robustness is desirable to allow self-supervised monocular trained depth estimation to address common mistakes made by the method, such as incorrect pose estimation and matching failures because of uniform textural details. We also show that our method can estimate pixel-wise depth uncertainties with the proposed discrete disparity volume~\cite{kendall2017end}. Depth uncertainty estimation is important for refining depth estimation~\cite{fu2018deep}, and in safety critical systems~\cite{kendall2017uncertainties}, allowing an agent to identify unknowns in an environment in order to reach optimal decisions. As a secondary contribution of this paper, we leverage recent advances in semantic segmentation network architectures that allow us to train larger models on a single GPU machine. Experimental results show that our novel approach produces the best self-supervised monocular depth estimation results for KITTI 2015 and Make3D. We also show in the experiments that our method is able to close the gap with self-supervised stereo trained and fully supervised depth estimators.

\section{Related Work}

Many computer vision and robotic systems that are used in navigation, localization and mapping rely on accurately understanding the 3D world around them~\cite{menze2015object, gupta2014learning, dudek2010computational, achtelik2009stereo}. Active sensors such as LiDAR, Time of Flight cameras, or Stereo/Multi camera rigs are often deployed in robotic and autonomous systems to estimate the depth of an image for understanding the agent's environment~\cite{dudek2010computational, achtelik2009stereo}.  Despite their wipe-spread adoption~\cite{rosique2019systematic}, these systems have several drawbacks~\cite{dudek2010computational}, including limited range, sensor noise, power consumption and cost. Instead of relying on these active sensor systems, recent advances leveraging fully supervised deep learning methods~\cite{eigen2014depth,eigen2015predicting, fu2018deep, guo2018learning, mayer2018makes} have made it possible to learn to predict depth from monocular RGB cameras~\cite{eigen2014depth, eigen2015predicting}. However, ground truth RGB-D data for supervised learning can be difficult to obtain, especially for every possible environment we wish our robotic agents to operate. To alleviate this requirement, many recent works have focused on developing self-supervised techniques to train monocular depth estimators using synchronised stereo image pairs~\cite{garg2016unsupervised, godard2017unsupervised, pillai2018superdepth}, monocular video~\cite{zhou2017unsupervised, monodepth2} or binocular video\cite{zhanst2018, monodepth2, luo2018every}. 

\subsection{Monocular Depth Estimation}

Depth estimation from a monocular image is an inherently ill-posed problem as pixels in the image can have multiple plausible depths. Nevertheless, methods based on supervised learning have been shown to mitigate this challenge and correctly estimate depth from colour input images~\cite{saxena2009make3d}. Eigen \etal~\cite{eigen2014depth} proposed the first method based on Deep Learning, which applies a multi-scale convolution neural network and a scale-invariant loss function to model local and global features within an image. Since then, fully supervised deep learning based methods have been continuously improved~\cite{fu2018deep, guo2018learning, mayer2018makes}. However these methods are limited by the availability of training data, which can be costly to obtain. While such issues can be mitigated with the use of synthetic training data~\cite{mayer2018makes}, simulated environments need to be modelled by human artists, limiting the amount of variation in the data set. 
To overcome fully supervised training set constraint, Garg \etal~\cite{garg2016unsupervised} propose a self-supervised framework, where instead of supervising using ground truth depth, a stereo photometric reprojection warping loss is used to implicitly learn depth.
This loss function is a pixel-based reconstruction loss that uses stereo pairs, 
where the right image of the pair is warped into the left using a differentiable image sampler~\cite{jaderberg2015spatial}. This loss function allows the deep learning model to implicitly recover the underlying depth for the input image. Expanding on this method, Godard \etal~\cite{godard2017unsupervised} add a left-right consistency loss term which helps to ensure consistency between the predicted depths from the left and right images of the stereo pair. While capable of training monocular depth estimators, these methods still rely on stereo-based training data which can still be difficult to acquire. This has motivated the development of self-supervised monocular trained depth estimators \cite{zhou2017unsupervised} which relax the requirement of synchronized stereo image pairs by jointly learning to predict depth and ego-motion with two separate networks, enabling the training of a monocular depth estimator using monocular video. To achieve this, the scene is assumed to be static (i.e., rigid), while the only motion is that of the camera. However, this causes degenerate behaviour in the depth estimator when this assumption is broken. To deal with this issue, the paper~\cite{zhou2017unsupervised} includes a predictive masking which learns to ignore regions that violates the rigidity assumptions. Vijayanarasimhan \etal~\cite{vijayanarasimhan2017sfm} propose a more complex motion model based on multiple motion masks, and GeoNet model~\cite{geonet2018} decomposes depth and optical flow to account for object motion within the image sequence. Self-supervised monocular trained methods have been further improved by constraining predicted depths to be consistent with surface normals~\cite{yang2017unsupervised}, using pre-computed instance-level segmentation masks~\cite{casser2018depth} and increasing the resolution of the input images~\cite{pillai2018superdepth}. Godard \etal~\cite{monodepth2} further close the performance gap between monocular and stereo-trained self-supervision with Monodepth2 which uses multi-scale estimation and a per-pixel minimum re-projection loss that better handles occlusions. We extend Monodepth2 with our proposed ideas, namely self-attention and discrete disparity volume.

\subsection{Self-attention}
\label{sec:self_attention}

Self-attention has improved the performance of natural language processing (NLP) systems by allowing a better handling of long-range dependencies between words~\cite{vaswani2017attention}, when compared with recurrent neural networks (RNN)~\cite{rumelhart1988learning}, long short term memory (LSTM)~\cite{lstm}, and convolutional neural nets (CNN)~\cite{lecun1998gradient}. This better performance can be explained by the fact that RNNs, LSTMs and CNNs can only process information in the local word neighbourhood, making these approaches insufficient for capturing long range dependencies in a sentence~\cite{vaswani2017attention}, which is essential in some tasks, like machine translation. Self-attention has been proposed in computer vision for improving Image Classification and Object Drection~\cite{bello2019attention, parmar2019stand}. Self-attention has also improved the performance of computer vision tasks such as semantic segmentation~\cite{yuan2018ocnet} by addressing more effectively the problem of segmenting visual classes in non-contiguous regions of the image, when compared with convolutional layers~\cite{Deeplab, pspnet, deeplabv3plus2018}, which can only process information in the local pixel neighbourhood. In fact, many of the recent improvements in semantic segmentation performance stem from improved contextual aggregation strategies (i.e., strategies that can process spatially non-contiguous image regions) such as the Pyramid Pooling Module (PPM) in PSPNet~\cite{pspnet}, and the Atrous Spatial Pyramid Pooling \cite{Deeplab}. In both of these methods, multiple scales of information are aggregated to improve the contextual representation by the network. Yuan \etal~\cite{yuan2018ocnet} further improve on this area with OCNet, which adds to a ResNet-101~\cite{he2016deep} backbone a self-attention module that learns to contextually represent groups of features with similar semantic similarity. Therefore, we hypothesise that such self-attention mechanisms can also improve depth prediction using monocular video because the correct context for the prediction of a pixel depth may be at a non-contiguous location that the standard convolutions cannot reach.

\vspace{-1pt}
\subsection{Discrete Disparity Volume}
\label{sec:DDV}

Kendall \etal~\cite{kendall2017end} propose to learn stereo matching in a supervised manner, by using a shared CNN encoder with a cost volume that is refined using 3D convolutions. 
Liu \etal~\cite{liu2019neural} investigate this idea further by training a model using monocular video with ground truth depth and poses. This paper~\cite{liu2019neural} relies on a depth probability volume (DPV) and a Bayesian filtering framework that refines outliers based on the uncertainty computed from the DPV. Fu \etal~\cite{fu2018deep} represent their ground-truth depth data as discrete bins, effectively forming a disparity volume for training. All methods above work in fully-supervised scenarios, showing advantages for depth estimation robustness and sharpness, allied with the possibility of estimating depth uncertainty. Such uncertainty estimation can be used by autonomous systems to improve decision making~\cite{kendall2017uncertainties} or to refine depth estimation~\cite{fu2018deep}. In this paper, we hypothesis that the extension of self-supervised monocular trained methods with a discrete disparity volume will provide the same advantages observed in fully-supervised models.




\begin{table*}[t!]
													
	\centering
	\resizebox{0.8\textwidth}{!}{
		\begin{tabular}{|l|c||c|c|c|c|c|c|c|}
			\hline
			Method                                                          & Train & \cellcolor{col1}Abs Rel & \cellcolor{col1}Sq Rel & \cellcolor{col1}RMSE & \cellcolor{col1}RMSE log & \cellcolor{col2}$\delta < 1.25 $ & \cellcolor{col2}$\delta < 1.25^{2}$ & \cellcolor{col2}$\delta < 1.25^{3}$ \\
			\hline
			Eigen \cite{eigen2014depth}                                     & D     & 0.203                   & 1.548                  & 6.307                & 0.282                    & 0.702                            & 0.890                               & 0.890                               \\
			Liu \cite{liu2015learning}                                      & D     & 0.201                   & 1.584                  & 6.471                & 0.273                    & 0.680                            & 0.898                               & 0.967                               \\
			Klodt \cite{klodt2018supervising}                               & D*M   & 0.166                   & 1.490                  & 5.998                & -                        & 0.778                            & 0.919                               & 0.966                               \\
			AdaDepth \cite{gandepth2018}                                    & D*    & 0.167                   & 1.257                  & 5.578                & 0.237                    & 0.771                            & 0.922                               & 0.971                               \\
			Kuznietsov \cite{kuznietsov2017semi}                            & DS    & 0.113                   & 0.741                  & 4.621                & 0.189                    & 0.862                            & 0.960                               & 0.986                               \\
			DVSO \cite{yang2018deep}                                        & D*S   & 0.097                   & 0.734                  & 4.442                & 0.187                    & 0.888                            & 0.958                               & 0.980                               \\
			SVSM FT \cite{singlestereo2018}                                 & DS    & \underline{0.094}       & \underline{0.626}      & 4.252                & 0.177                    & 0.891                            & 0.965                               & 0.984                               \\
			Guo \cite{guo2018learning}                                      & DS    & 0.096                   & 0.641                  & \underline{4.095}    & \underline{0.168}        & \underline{0.892}                & \underline{0.967}                   & \underline{0.986}                   \\
			DORN \cite{fu2018deep}                                          & D     & \textbf{0.072}          & \textbf{0.307}         & \textbf{2.727}       & \textbf{0.120}           & \textbf{0.932}                   & \textbf{0.984}                      & \textbf{0.994}                      \\ 
																																							
			\arrayrulecolor{black}\hline
																																							
			Zhou \cite{zhou2017unsupervised}\textdagger                     & M     & 0.183                   & 1.595                  & 6.709                & 0.270                    & 0.734                            & 0.902                               & 0.959                               \\
			Yang \cite{yang2017unsupervised}                                & M     & 0.182                   & 1.481                  & 6.501                & 0.267                    & 0.725                            & 0.906                               & 0.963                               \\
			Mahjourian \cite{mahjourian2018unsupervised}                    & M     & 0.163                   & 1.240                  & 6.220                & 0.250                    & 0.762                            & 0.916                               & 0.968                               \\
																																							
			GeoNet \cite{geonet2018}\textdagger                             & M     & 0.149                   & 1.060                  & 5.567                & 0.226                    & 0.796                            & 0.935                               & 0.975                               \\
			DDVO \cite{wang2017learning}                                    & M     & 0.151                   & 1.257                  & 5.583                & 0.228                    & 0.810                            & 0.936                               & 0.974                               \\
			DF-Net \cite{zou2018df}                                         & M     & 0.150                   & 1.124                  & 5.507                & 0.223                    & 0.806                            & 0.933                               & 0.973                               \\
			LEGO \cite{yang2018lego}                                        & M     & 0.162                   & 1.352                  & 6.276                & 0.252                    & -                                & -                                   & -                                   \\
			Ranjan \cite{ranjan2018adversarial}                             & M     & 0.148                   & 1.149                  & 5.464                & 0.226                    & 0.815                            & 0.935                               & 0.973                               \\
			EPC++ \cite{luo2018every}                                       & M     & 0.141                   & 1.029                  & 5.350                & 0.216                    & 0.816                            & 0.941                               & 0.976                               \\
			Struct2depth `(M)' \cite{casser2018depth}                       & M     & 0.141                   & 1.026      & 5.291                & 0.215                    & 0.816                            & 0.945                               & 0.979                   \\																		
																																							        
			Monodepth2 \cite{monodepth2}                                    & M     & \underline{0.115}                   & 0.903                  & 4.863                & 0.193                    & 0.877                            & 0.959                               & 0.981                               \\ 
			Monodepth2 (1024 $\times$ 320)\cite{monodepth2}                 & M     & \underline{0.115}          & \underline{0.882}         & \underline{4.701}       & \underline{0.190}           & \underline{0.879}                   & \underline{0.961}                      & \textbf{0.982}                      \\

			\rowcolor{gray!30} {\bf Ours}                                                     & M     & {\bf 0.106}             & {\bf 0.861}            & {\bf 4.699}          & {\bf 0.185}              & {\bf 0.889}                      & {\bf 0.962}                         & {\bf 0.982}                         \\ 
																																							
			\arrayrulecolor{black}\hline
																																							
			Garg \cite{garg2016unsupervised}\textdagger                     & S     & 0.152                   & 1.226                  & 5.849                & 0.246                    & 0.784                            & 0.921                               & 0.967                               \\
			Monodepth R50 \cite{godard2017unsupervised}\textdagger          & S     & 0.133                   & 1.142                  & 5.533                & 0.230                    & 0.830                            & 0.936                               & 0.970                               \\
			StrAT \cite{mehta2018structured}                                & S     & 0.128                   & 1.019                  & 5.403                & 0.227                    & 0.827                            & 0.935                               & 0.971                               \\
			3Net  (R50) \cite{poggi20183net}                                & S     & 0.129                   & 0.996                  & 5.281                & 0.223                    & 0.831                            & 0.939                               & 0.974                               \\
			3Net (VGG)  \cite{poggi20183net}                                & S     & 0.119                   & 1.201                  & 5.888                & 0.208                    & 0.844                            & 0.941                               & \textbf{0.978}                      \\
			SuperDepth + pp \cite{pillai2018superdepth} (1024 $\times$ 382) & S     & {0.112}       & {0.875}      & \underline{4.958}       & \underline{0.207}           & \underline{0.852}                & {0.947}                   & \underline{0.977}                   \\
																																							
			Monodepth2 \cite{monodepth2}                                    & S     & \underline{0.109}          & \underline{0.873}         & {4.960}    & {0.209}        & {0.864}                   & \underline{0.948}                      & 0.975                               \\ 
																																							
			Monodepth2 (1024 $\times$ 320)\cite{monodepth2}                 & S     & \textbf{0.107}          & \textbf{0.849}         & \textbf{4.764}       & \textbf{0.201}           & \textbf{0.874}                   & \textbf{0.953}                      & \underline{0.977}                   \\ 
																																							
			\arrayrulecolor{black}\hline
																																							
			UnDeepVO \cite{li2017undeepvo}                                  & MS    & 0.183                   & 1.730                  & 6.57                 & 0.268                    & -                                & -                                   & -                                   \\
			Zhan FullNYU \cite{zhanst2018}                                  & D*MS  & 0.135                   & 1.132                  & 5.585                & 0.229                    & 0.820                            & 0.933                               & 0.971                               \\
			EPC++ \cite{luo2018every}                                       & MS    & \underline{0.128}                   & {0.935}      & {5.011}    & {0.209}        & 0.831                            & {0.945}                   & \underline{0.979}                      \\
			Monodepth2\cite{monodepth2}                                     & MS    & \textbf{0.106}          & \underline{0.818}         & \underline{4.750}       & \underline{0.196}           & \underline{0.874}                   & \underline{0.957}                      & \underline{0.979}                      \\
																																							
			Monodepth2(1024 $\times$ 320)\cite{monodepth2}                  & MS    & \textbf{0.106}          & \textbf{0.806}         & \textbf{4.630}       & \textbf{0.193}           & \textbf{0.876}                   & \textbf{0.958}                      & \textbf{0.980}                      \\
																																							    
			\arrayrulecolor{black}\hline
																																							
		\end{tabular}
	}
	\caption{\textbf{Quantitative results.} Comparison of existing methods to our own on the KITTI 2015  \cite{Geiger2012CVPR} using the Eigen split \cite{eigen2015predicting}. The Best results are presented in {\bf bold} for each category, with second best results \underline{underlined}. The supervision level for each method is presented in the \emph{Train} column with; D -- Depth Supervision, D* -- Auxiliary depth supervision, S -- Self-supervised stereo supervision, M -- Self-supervised mono supervision. Results are presented without any post-processing \cite{godard2017unsupervised}, unless marked with -- + pp. If newer results are available on github, these are marked with -- \textdagger. Non-Standard resolutions are documented along with the method name.  Metrics indicated by red: \emph{lower is better}, Metrics indicated by blue: \emph{higher is better}}
    \label{tab:kitti_eigen}
\end{table*}


\vspace{-3pt}
\section{Methods}
\label{sec:methods}
\vspace{-1pt}
In the presentation of our proposed model for self-supervised monocular trained depth estimation, we focus on showing the importance of the main contributions of this paper, namely self-attention and discrete disparity volume.
We use as baseline, the Monodepth2 model~\cite{monodepth2} based on a UNet architecture~\cite{ronneberger2015u}.

\subsection{Model}

We represent the RGB image with $\mathbf{I}:\Omega \rightarrow \mathbb R^3$, where $\Omega$ denotes the image lattice of height $H$ and width $W$.  The first stage of the model, depicted in Fig.~\ref{fig:arch}, is the ResNet-101 encoder, which forms $\mathbf{X} = resnet_{\theta}(\mathbf{I}_t)$, with $\mathbf{X}:\Omega_{1/8} \rightarrow \mathbb R^{M}$, $M$ denoting the number of channels at the output of the ResNet, and $\Omega_{1/8}$ representing the low-resolution lattice at $(1/8)^{th}$ of its initial size in $\Omega$. The ResNet output is then used by the self-attention module~\cite{non-local}, which first forms the query, key and value results, represented by:
\begin{equation}
\begin{split}
    f(\mathbf{X}(\omega)) = & \mathbf{W}_f\mathbf{X}(\omega), \\ g(\mathbf{X}(\omega)) = & \mathbf{W}_g\mathbf{X}(\omega), \\
    h(\mathbf{X}(\omega)) = & \mathbf{W}_h\mathbf{X}(\omega),
    \label{eq:self_attention_defintion_key_query_value}
\end{split}
\end{equation}
respectively, with
$\mathbf{W}_f,\mathbf{W}_g,\mathbf{W}_h \in \mathbb R^{N \times M}$.  The query and key values are then combined with
\begin{equation}
    \mathbf{S}_{\omega} = softmax(f(\mathbf{X}(\omega))^T g(\mathbf{X}) ),
    \label{eq:self_attention_query_times_key}
\end{equation}
where $\mathbf{S}_{\omega}: \Omega_{1/8} \rightarrow [0,1]$, and we abuse the notation by representing $g(\mathbf{X})$ as a tensor of size $N \times H/8 \times W/8$.  The self-attention map is then built by the multiplication of value and $\mathbf{S}_{\omega}$ in~\eqref{eq:self_attention_query_times_key}, with:
\begin{equation}
\mathbf{A}(\omega) = \sum_{\tilde{\omega} \in \Omega_{1/8}} h(\mathbf{X}(\tilde{\omega})) \times \mathbf{S}_{\omega}(\tilde{\omega}), 
\label{eq:self_attention}
\end{equation}
with $\mathbf{A}:\Omega_{1/8} \rightarrow \mathbb R^N$.

The low-resolution discrete disparity volume (DDV) is denoted by $\mathbf{D}_{1/8}(\omega) = conv_{3 \times 3}(\mathbf{A}(\omega))$, with $\mathbf{D}_{1/8}:\Omega_{1/8} \rightarrow \mathbb R^K$ ($K$ denotes the number of discretized disparity values), and $conv_{3 \times 3}(.)$ denoting a convolutional layer with filters of size $3 \times 3$.  The low resolution disparity map is then computed with
\begin{equation}
	\sigma(\mathbf{D}_{1/8}(\omega)) = \sum_{k=1}^{K} softmax(\mathbf{D}_{1/8}(\omega)[k]) \times disparity(k),
	\label{eq:DDV_low_res}
\end{equation}
where $softmax(\mathbf{D}_{1/8}(\omega)[k])$ is the softmax result of the $k^{th}$ output from $\mathbf{D}_{1/8}$, and $disparity(k)$ holds the disparity value for $k$.  Given the ambiguous results produced by these low-resolution disparity maps, we follow the multi-scale strategy proposed by Godard \etal~\cite{monodepth2}.
The low resolution map from~\eqref{eq:DDV_low_res} is the first step of the multi-scale decoder that consists of three additional stages of upconv operators (i.e., nearest upsample + convolution) that receive skip connections from the ResNet encoder for the respective resolutions, as shown in Fig.~\ref{fig:arch}. These skip connections between encoding layers and associated decoding layers are known to retain high-level information in the final depth output.  At each resolution, we form a new DDV, which is used to compute the disparity map at that particular resolution.  The resolutions considered are (1/8), (1/4), (1/2), and (1/1) of the original resolution, respectively represented by $\sigma(\mathbf{D}_{1/8})$, $\sigma(\mathbf{D}_{1/4})$, $\sigma(\mathbf{D}_{1/2})$, and $\sigma(\mathbf{D}_{1/1})$.

Another essential part of our model is the pose estimator~\cite{zhou2017unsupervised}, which takes two images recorded at two different time steps, and returns the relative transformation, as in 
\begin{equation}
    \mathbf{T}_{t \rightarrow t'} = p_{\phi}(\mathbf{I}_t,\mathbf{I}_{t'}),
    \label{eq:pose}
\end{equation}
where $\mathbf{T}_{t \rightarrow t'}$ denotes the transformation matrix between images recorded at time steps $t$ and $t'$, and $p_{\phi}(.)$ is the pose estimator, consisting of a deep learning model parameterised by $\phi$.

\subsection{Training and Inference}

The training is based on the minimum per-pixel photometric re-projection error~\cite{monodepth2} between the source image $\mathbf{I}_{t'}$ and the target image $\mathbf{I}_{t}$, using the relative pose $\mathbf{T}_{t \rightarrow t'}$ defined in~\eqref{eq:pose}.  The pixel-wise error is defined by
\begin{equation}
    \ell_p = \frac{1}{|\mathcal{S}|}\sum_{s \in \mathcal{S}} \left ( \min_{t'} \mu^{(s)} \times pe(\mathbf{I}_t,\mathbf{I}^{(s)}_{t \rightarrow t'}) \right ),
    \label{eq:photo_rep_loss}
\end{equation}
where $pe(.)$ denotes the photometric reconstruction error, $\mathcal{S}=\{ \frac{1}{8},\frac{1}{4},\frac{1}{2},\frac{1}{1}  \}$ is the set of the resolutions available for the disparity map, defined in~\eqref{eq:DDV_low_res},  $t' \in \{ t-1, t+1\}$, indicating that we use two frames that are temporally adjacent to $\mathbf{I}_t$ as its source frames~\cite{monodepth2}, and $\mu^{(s)}$ is a binary mask that filters out stationary points (see more details below in Eq.\ref{eq:automasking})~\cite{monodepth2}.  The re-projected image in~\eqref{eq:photo_rep_loss} is defined by
\begin{equation}
    \mathbf{I}^{(s)}_{t \rightarrow t'} = \mathbf{I}_{t'} \big < proj(\sigma(\mathbf{D}^{(s)}_t), \mathbf{T}_{t \rightarrow t'} , \mathbf{K}) \big >,
\end{equation}
where $proj(.)$ represents the 2D coordinates of the projected depths $\mathbf{D}_t$ in $\mathbf{I}_{t'}$, $\big <. \big >$ is the sampling operator, and $\sigma(\mathbf{D}^{(s)}_t)$ is defined in~\eqref{eq:DDV_low_res}.
Similarly to~\cite{monodepth2}, the pre-computed intrinsics $\mathbf{K}$ of all images are identical, and we use bi-linear sampling to sample the source images and
\begin{equation}
	\quad pe(\mathbf{I}_t, \mathbf{I}^{(s)}_{t^\prime}) = \frac{\alpha}{2} (1 - \mathrm{SSIM}(\mathbf{I}_t, \mathbf{I}^{(s)}_{t^\prime})) + (1 - \alpha) \|\mathbf{I}_t - \mathbf{I}^{(s)}_{t^\prime}\|_1,
\end{equation}
where $\alpha = 0.85$.  
Following \cite{godard2017unsupervised} we use an edge-aware smoothness regularisation term to improve the predictions around object boundaries: 
\begin{eqnarray}
	\ell_s &=& \left | \partial_x d^*_t   \right | e^{-\left | \partial_x \mathbf{I}_t \right |} + \left | \partial_y d^*_t   \right | e^{-\left | \partial_y \mathbf{I}_t \right |},
    \label{eq:smoothness}
\end{eqnarray} 
where $d^*_t = d_t / \overline{d_t}$ is the mean-normalized inverse depth from \cite{wang2017learning} to discourage shrinking of the estimated depth.
The auto-masking of stationary points~\cite{monodepth2} in~\eqref{eq:photo_rep_loss} is necessary because the assumptions of a moving camera and a static scene are not always met in self-supervised monocular trained depth estimation methods~\cite{monodepth2}. This masking filters out pixels that remain with the same appearance between two frames in a sequence, and is achieved with a binary mask defined as
\begin{equation}
    \mu^{(s)} = \big [ \min_{t'}pe(\mathbf{I}_t,\mathbf{I}^{(s)}_{t' \rightarrow t}) < \min_{t'}pe(\mathbf{I}_t,\mathbf{I}_{t'}) \big ], 
    \label{eq:automasking}
\end{equation}
where $[.]$ represents the Iverson bracket. The binary mask $\mu$ in~\eqref{eq:automasking} masks the loss in~\eqref{eq:photo_rep_loss} to only include the pixels where the re-projection error of $\mathbf{I}^{(s)}_{t' \rightarrow t}$ is lower than the error of the un-warped image $\mathbf{I}_{t'}$, indicating that the visual object is moving relative to the camera. The final loss is computed as the weighted sum of the per-pixel minimum reprojection loss in~\eqref{eq:photo_rep_loss} and  smoothness term in~\eqref{eq:smoothness},
\begin{equation}
\ell = \ell_{p} + \lambda\ell_s 
\label{eq:final-loss}
\end{equation}
where $\lambda$ is the weighting for the smoothness regularisation term. Both the pose model and depth model are trained jointly using this photometric reprojection error. Inference is achieved by taking a test image at the input of the model and producing the high-resolution disparity map $\sigma(\mathbf{D}_{1/1})$.

\begin{figure*}
	\centering
    \resizebox{\textwidth}{!}{
		\newcommand{\turnheightnew}{0.195\columnwidth}

\centering

\begin{tabular}{@{\hskip 2mm}c@{\hskip 2mm}c@{\hskip 2mm}c@{\hskip 2mm}c@{}}

{\rotatebox{90}{\hspace{4mm}Input}} &
\includegraphics[height=\turnheightnew]{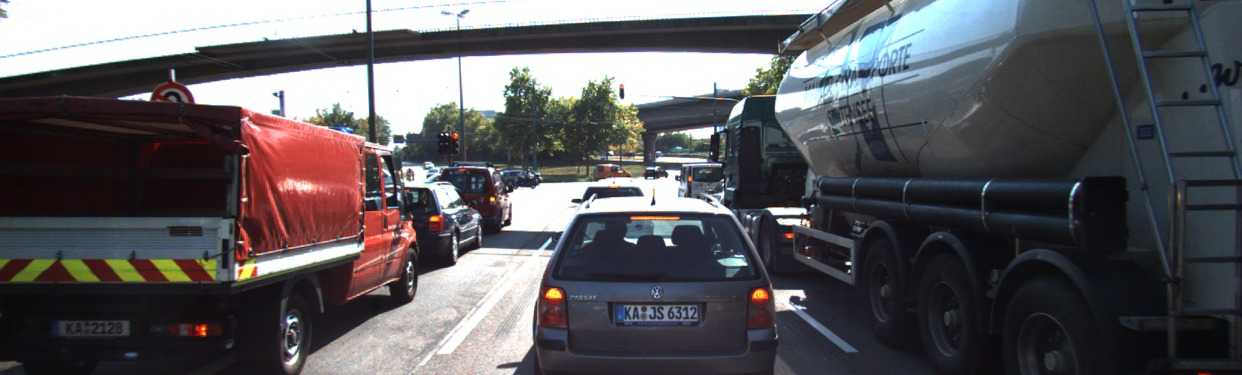} &
\includegraphics[height=\turnheightnew]{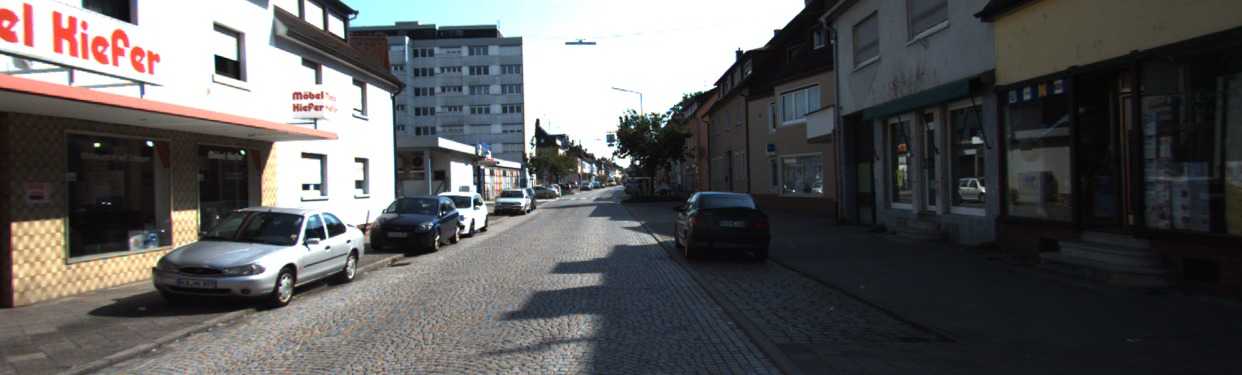} &
\includegraphics[height=\turnheightnew]{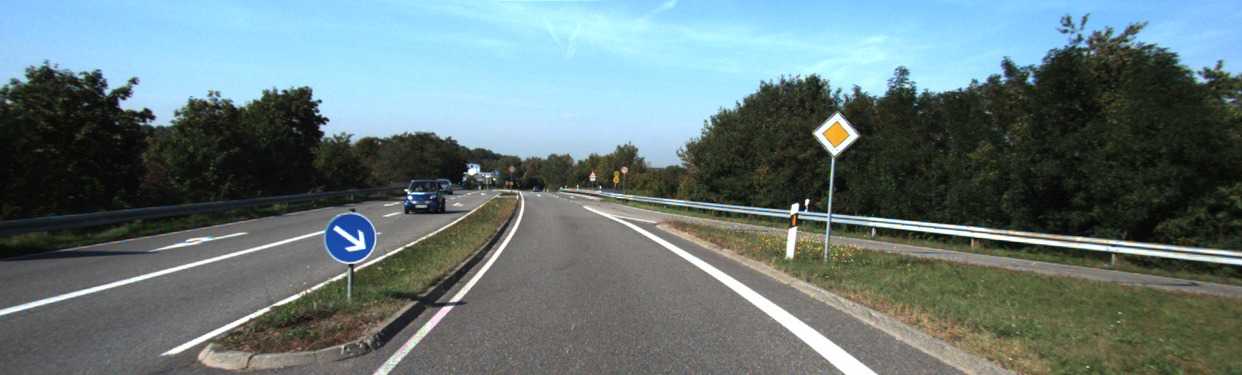} \\

{\rotatebox{90}{\scriptsize \hspace{0mm}\textbf{MD2 MS~\cite{monodepth2}}}} &
\includegraphics[height=\turnheightnew]{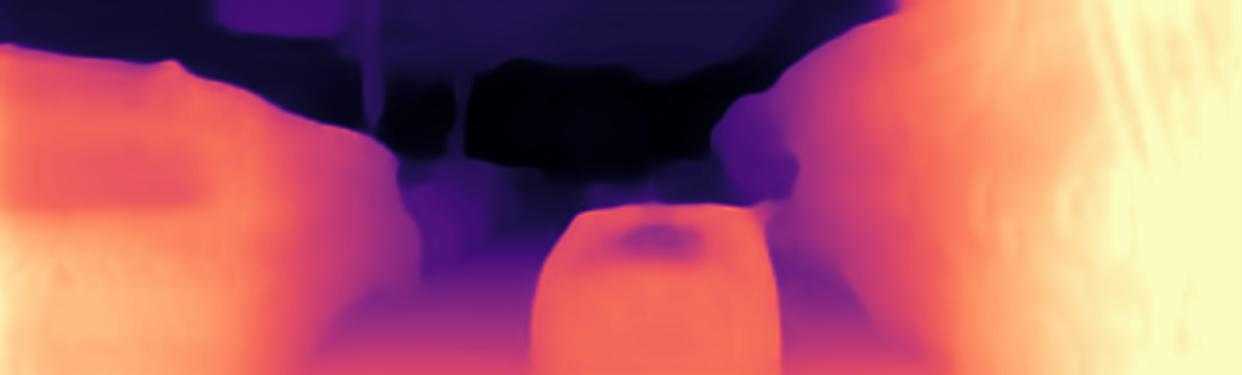} &
\includegraphics[height=\turnheightnew]{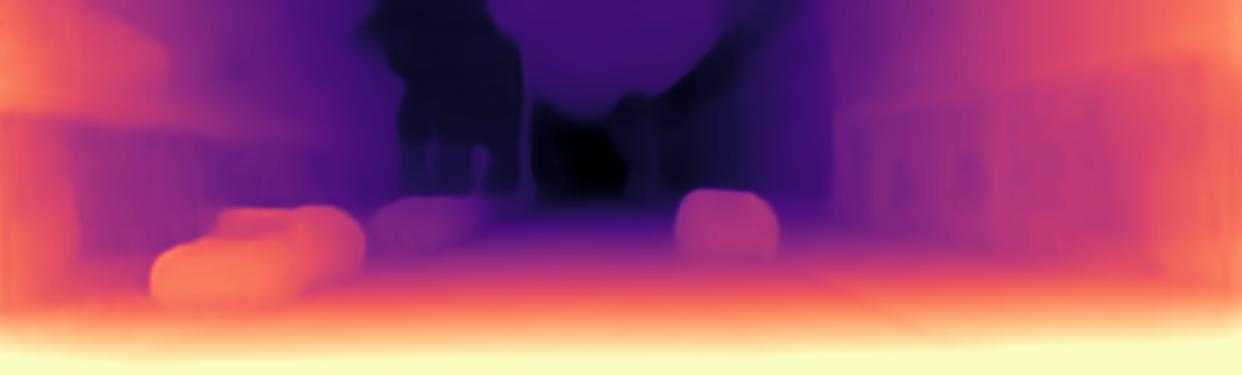} &
\includegraphics[height=\turnheightnew]{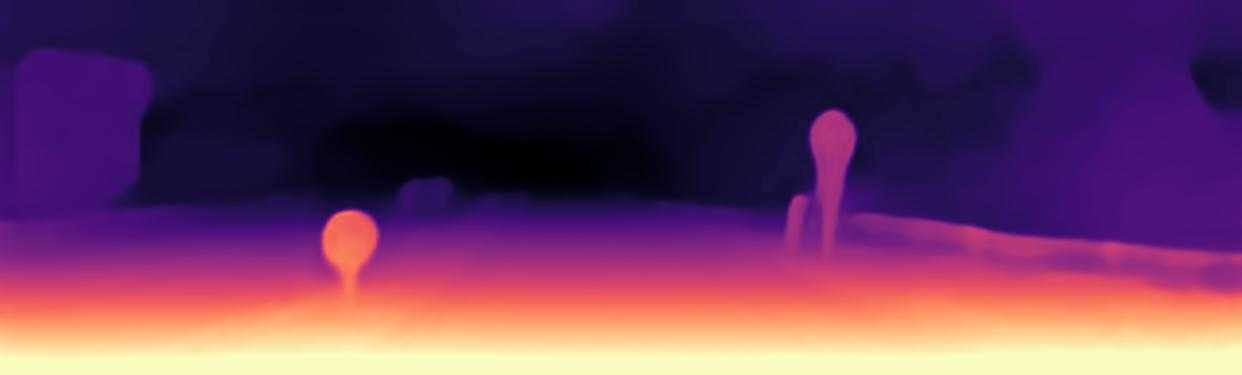} \\

{\rotatebox{90}{\scriptsize \hspace{0mm}\textbf{MD2 M~\cite{monodepth2}}}} &
\includegraphics[height=\turnheightnew]{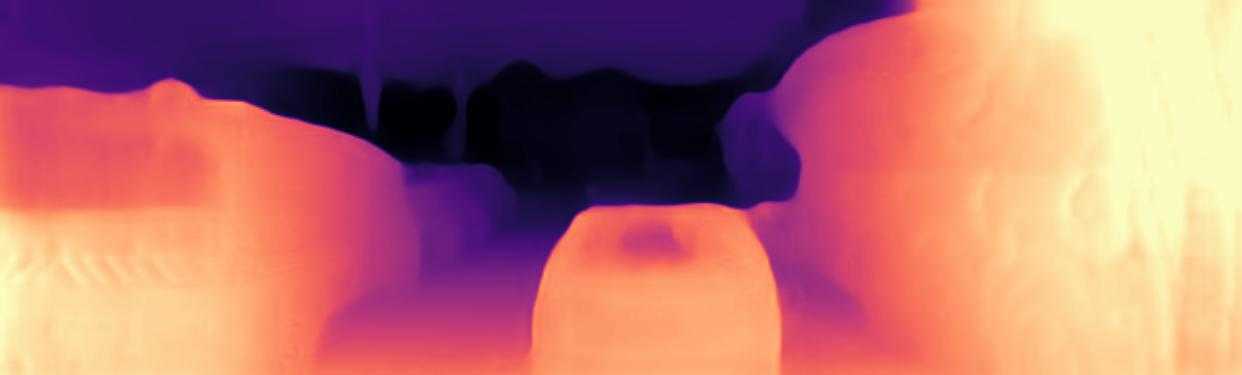} &
\includegraphics[height=\turnheightnew]{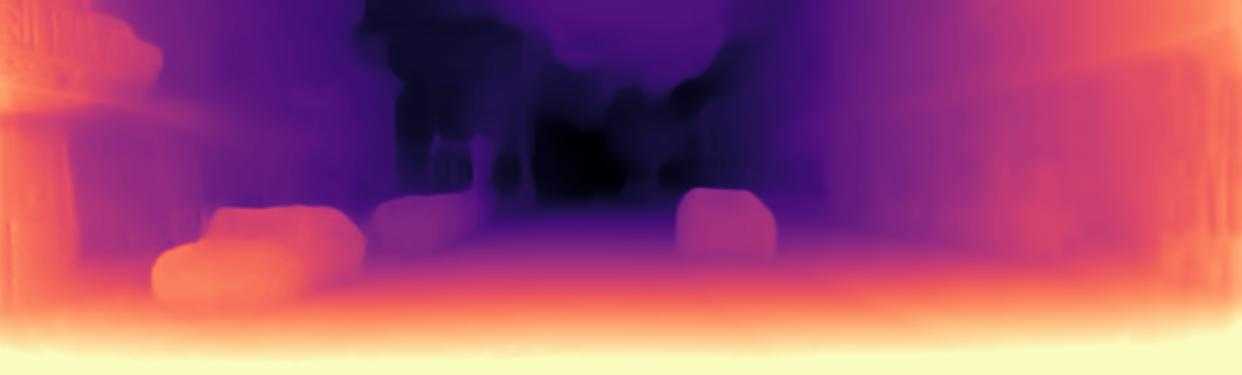} &
\includegraphics[height=\turnheightnew]{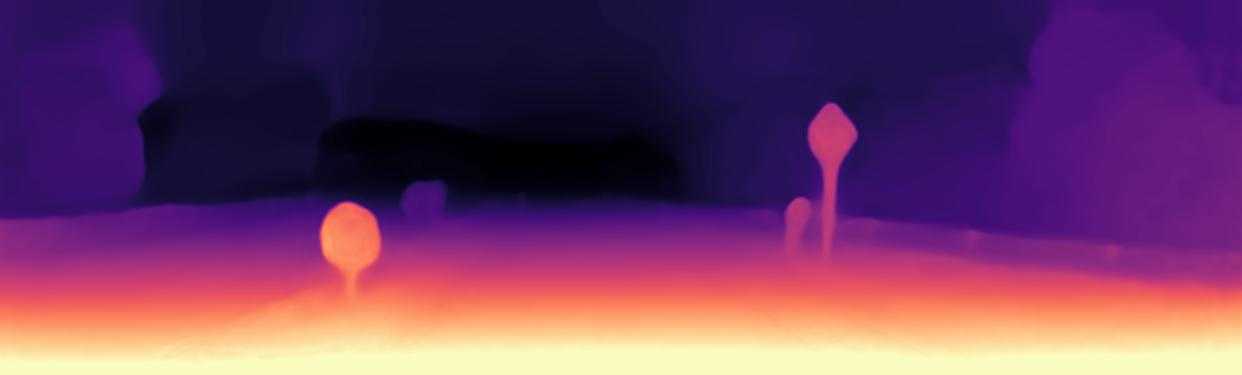} \\

{\rotatebox{90}{\scriptsize \hspace{5mm}\textbf{Ours}}} &
\includegraphics[height=\turnheightnew]{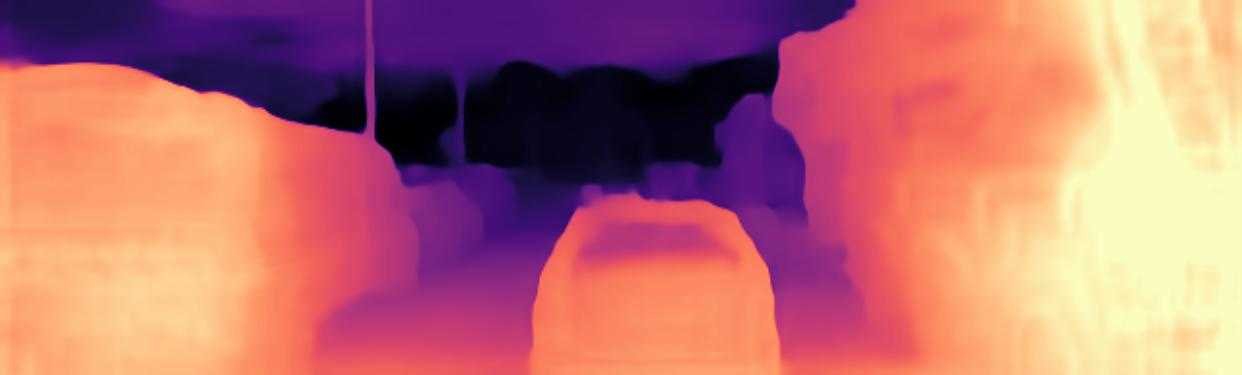} &
\includegraphics[height=\turnheightnew]{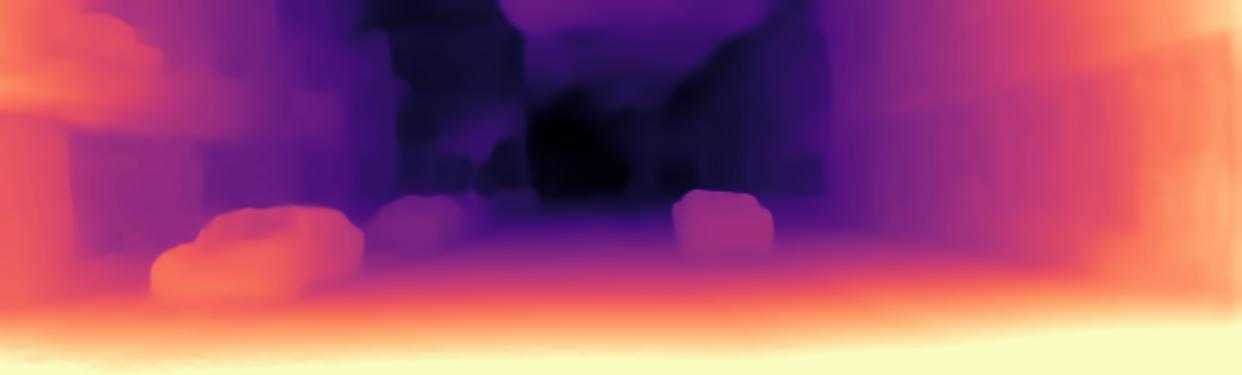} &
\includegraphics[height=\turnheightnew]{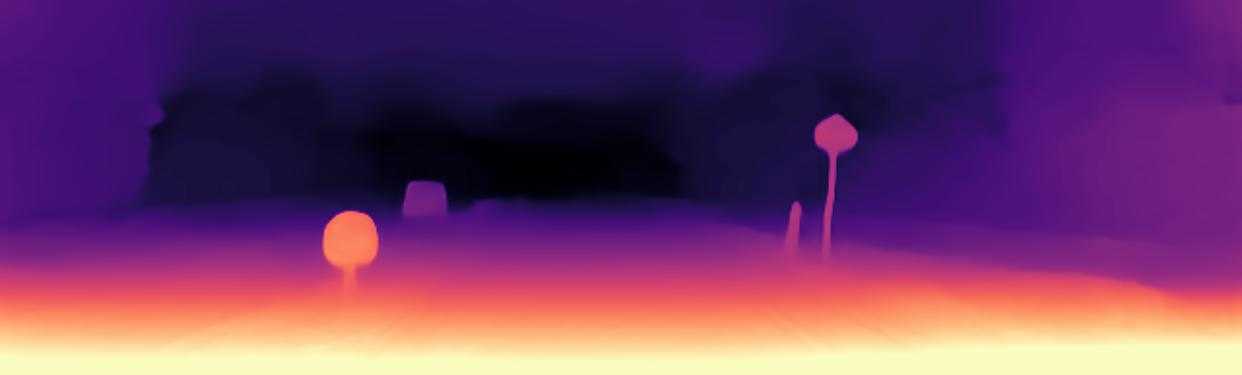} \\

{\rotatebox{90}{\hspace{4mm}Input}} &
\includegraphics[height=\turnheightnew]{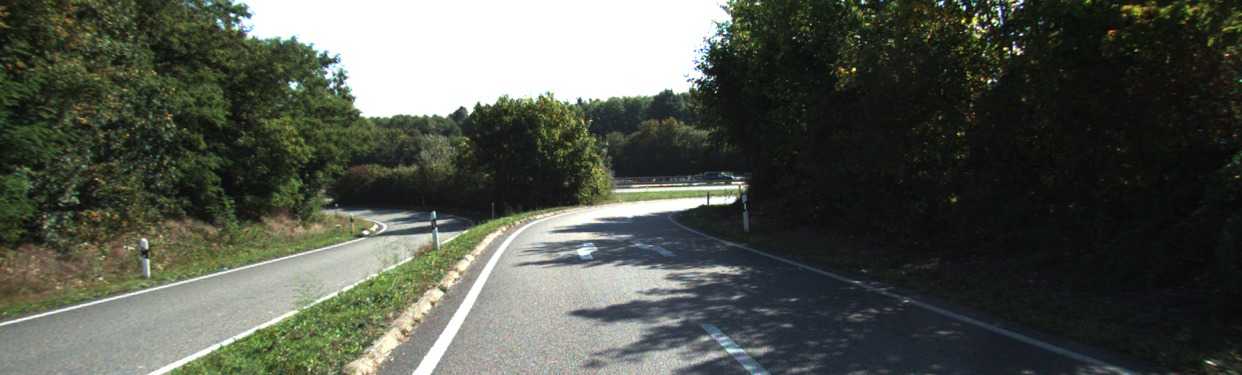} &
\includegraphics[height=\turnheightnew]{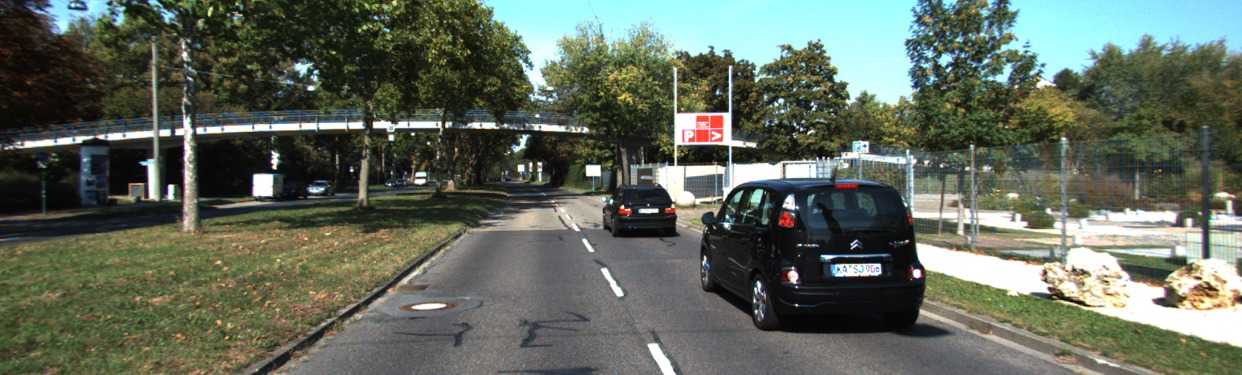} &
\includegraphics[height=\turnheightnew]{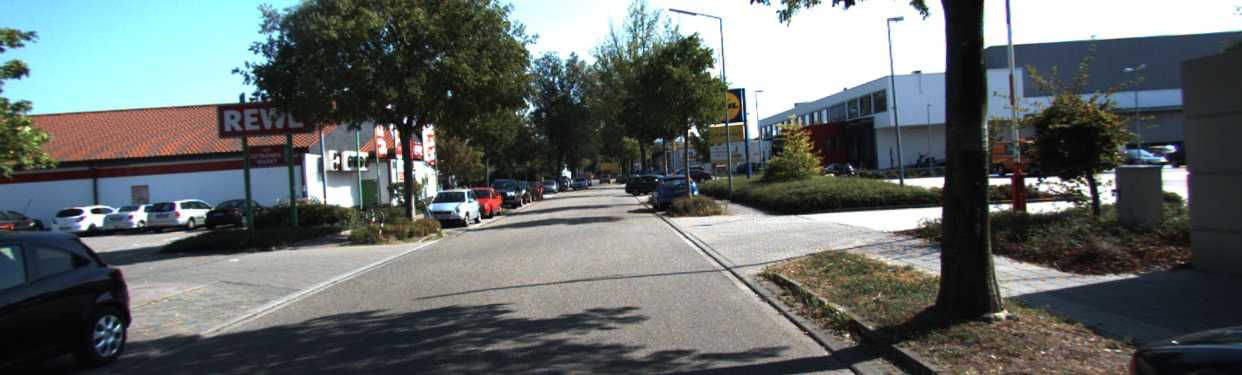} \\

{\rotatebox{90}{\scriptsize \hspace{0mm}\textbf{MD2 MS~\cite{monodepth2}}}} &
\includegraphics[height=\turnheightnew]{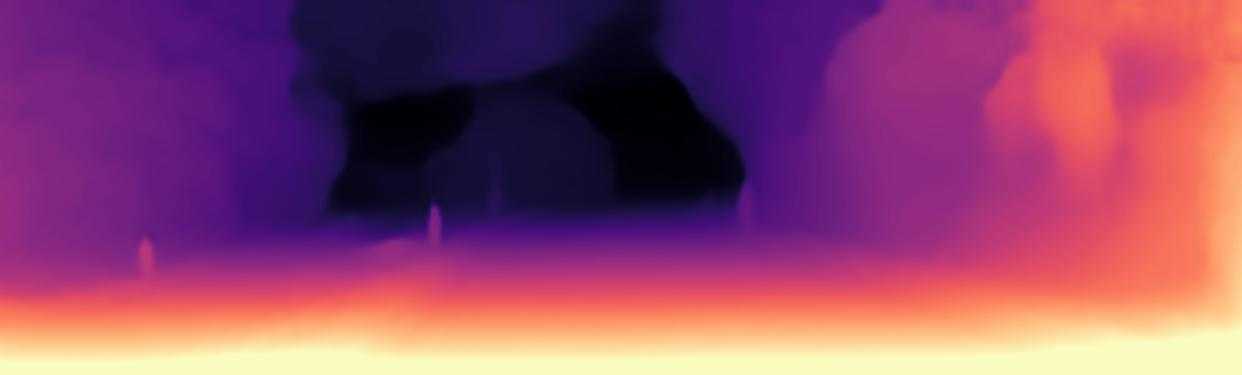} &
\includegraphics[height=\turnheightnew]{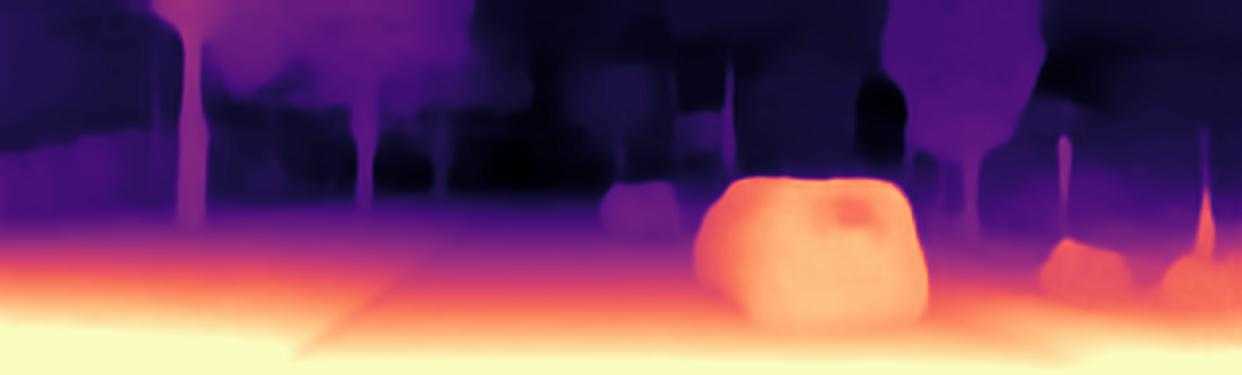} &
\includegraphics[height=\turnheightnew]{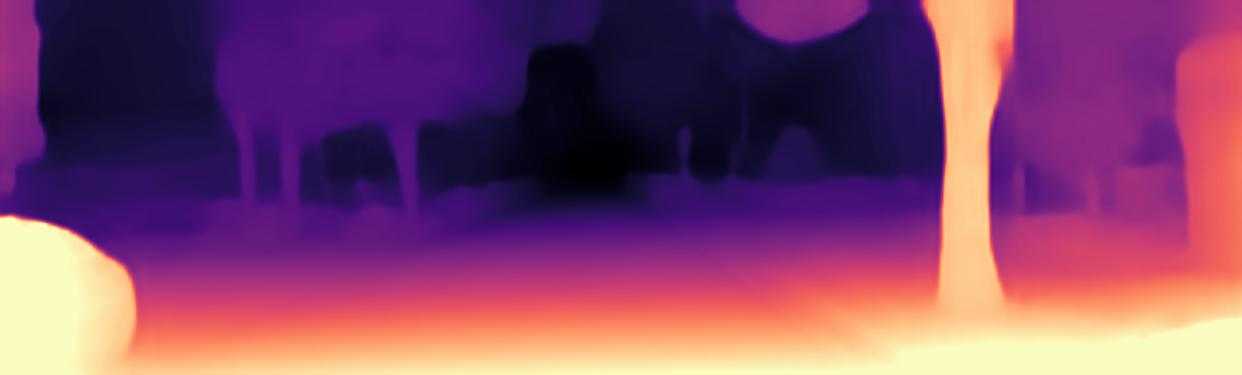} \\

{\rotatebox{90}{\scriptsize \hspace{0mm}\textbf{MD2 M~\cite{monodepth2}}}} &
\includegraphics[height=\turnheightnew]{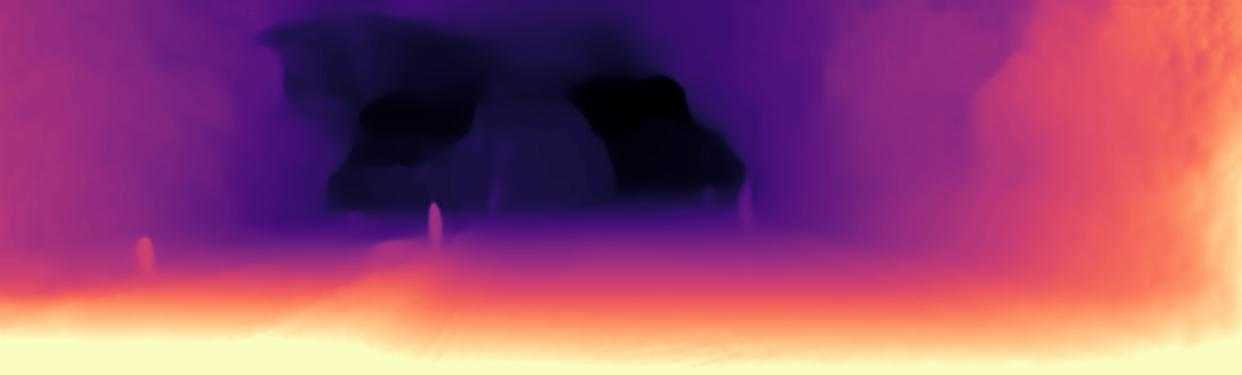} &
\includegraphics[height=\turnheightnew]{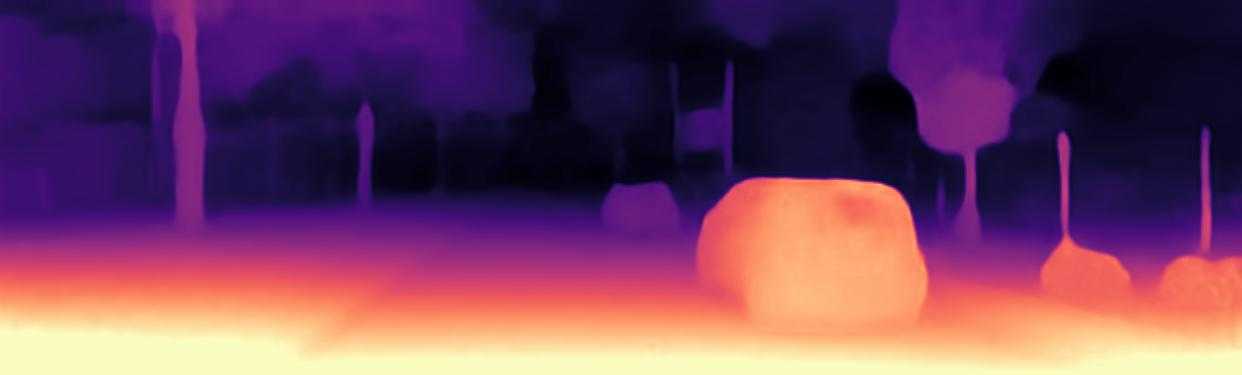} &
\includegraphics[height=\turnheightnew]{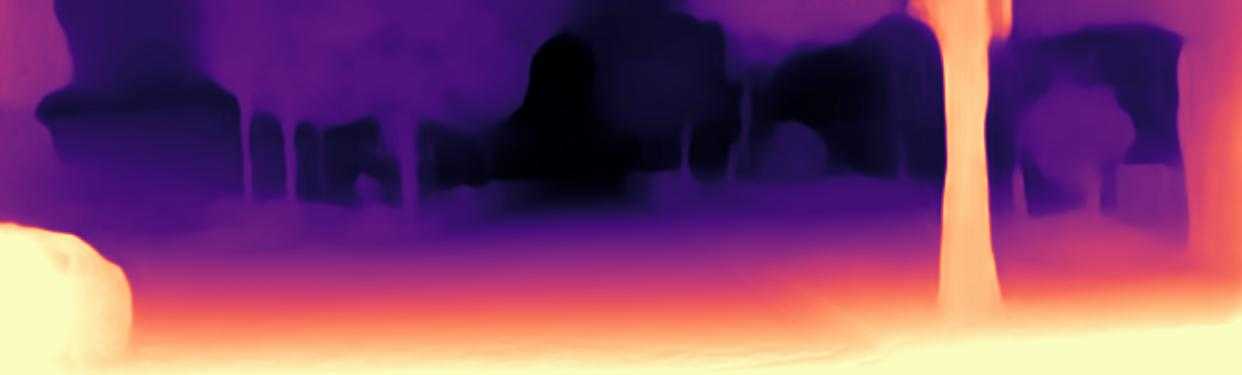} \\

{\rotatebox{90}{\scriptsize \hspace{5mm}\textbf{Ours}}} &
\includegraphics[height=\turnheightnew]{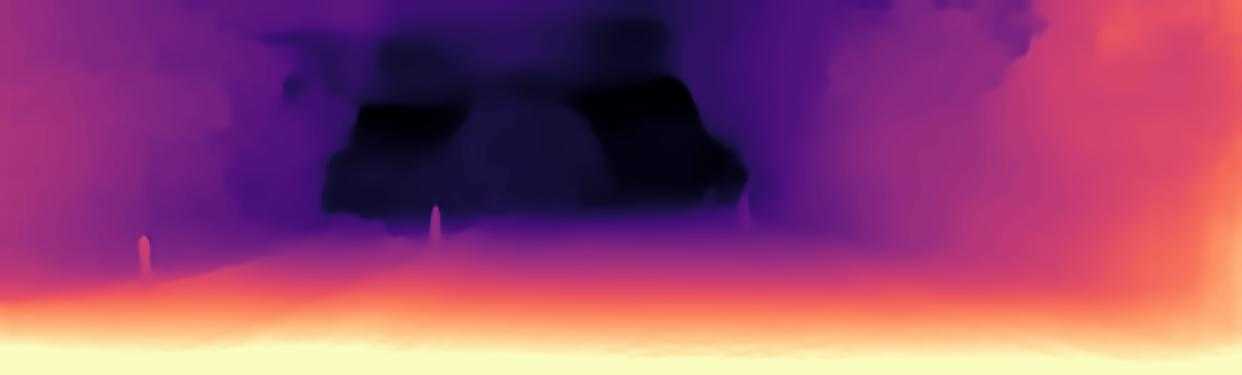} &
\includegraphics[height=\turnheightnew]{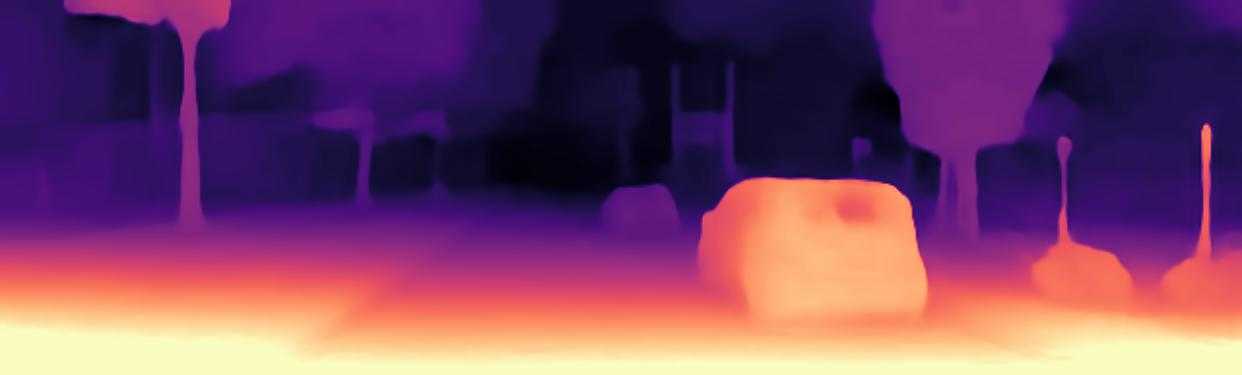} &
\includegraphics[height=\turnheightnew]{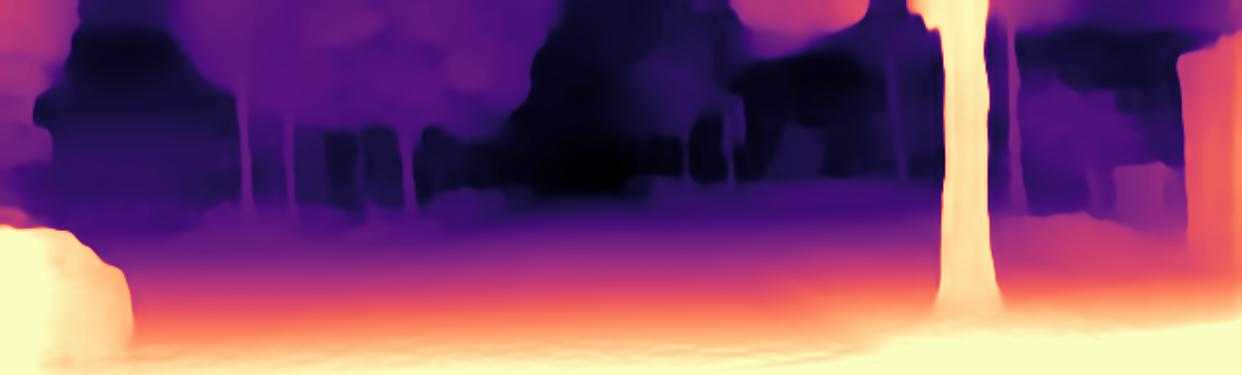} \\

\end{tabular}

    }
   \caption{{\bf Qualitative results on the KITTI Eigen split~\cite{eigen2015predicting} test set.} Our models perform better on thinner objects such as trees, signs and bollards, as well as being better at delineating difficult object boundaries.}
    \vspace*{-10pt}
	\label{fig:qual-results}
\end{figure*}

\vspace{-2pt}
\section{Experiments}
\label{sec:experiments}
\vspace{-1pt}


We train and evaluate our method using the KITTI 2015 stereo data set~\cite{Geiger2012CVPR}. We also evaluate our method on the Make3D data set~\cite{saxena2009make3d} using our model trained on KITTI 2015.
We use the split and evaluation of Eigen \etal~\cite{eigen2015predicting}, and following previous works \cite{zhou2017unsupervised,monodepth2}, we remove static frames before training and only evaluate depths up to a fixed range of 80m~\cite{eigen2015predicting, garg2016unsupervised, godard2017unsupervised, monodepth2}. As with~\cite{monodepth2}, this results in 39,810 monocular training sequences, consisting of sequences of three frames, with 4,424 validation sequences. 
As our baseline model, we use Monodepth2~\cite{monodepth2}, but we replace the original ResNet-18 by a ResNet-101 that has higher capacity, but requires more memory. To address this memory issue, we use the inplace activated batch normalisation~\cite{rota2018place}, which fuses the batch normalization layer and the activation functions to reach up to 50\% memory savings.
As self-supervised monocular trained depth estimators do not contain scale information, we use the per-image median ground truth scaling~\cite{zhou2017unsupervised, monodepth2}.
Following architecture best practices from the Semantic Segmentation community, we adopt the \emph{atrous convolution}~\cite{chen2017rethinking}, also known as the \emph{dilated convolution}, in the last two convolutional blocks of the ResNet-101 encoder~\cite{pspnet, yuan2018ocnet, chen2017rethinking, deeplabv3plus2018} with dilation rates of 2 and 4, respectively. This has been shown to significantly improve multi-scale encoding by increasing the models field-of-view \cite{chen2017rethinking}. The results for the quantitative analysis are shown in Sec.~\ref{sec:kitti}. We also present an ablation study comparing the effects of the our different contributions in Sec.~\ref{sec:ablation}. Final models are selected using the lowest absolute relative error metric on the validation set. 
\vspace{-4pt}
\subsection{Implementation Details}
\vspace{-1pt}
Our system is trained using the PyTorch library \cite{pytorch}, with models trained on a single Nvidia 2080Ti for 20 epochs. We jointly optimize both our pose and depth networks with the Adam Optimizer~\cite{adamsolver} with $\beta_1 = 0.9$, $\beta_2 = 0.999$ and a learning rate of $1e^{-4}$. We use a single learning rate decay to $lr=1e^{-5}$ after 15 epochs. As with previous papers~\cite{monodepth2}, our ResNet encoders use pre-trained ImageNet~\cite{russakovsky2015imagenet} weights as this has been show to reduce training time and improve overall accuracy of the predicted depths. All models are trained using the following data augmentations with 50\% probability; Horizontal flips, random contrast ($\pm0.2$), saturation ($\pm0.2$), hue jitter ($\pm0.1$) and brightness ($\pm0.2$). Crucially, augmentations are only performed on the images input into the depth and pose network and the loss in~\eqref{eq:final-loss} is computed using the original ground truth images, with the smoothness term set to $\lambda=1e^{-3}$. Image resolution is set to $640 \times 192$ pixels. 

\vspace{-1pt}
\subsection{KITTI Results}
\label{sec:kitti}
\begin{figure}
	\centering
	\newcommand{\imlabel}[2]{\includegraphics[width=0.49\columnwidth]{#1}%
\raisebox{2pt}{\makebox[-2pt][r]{\footnotesize #2}}}
\newcommand{\turnheightnew}{0.195\columnwidth}
\centering

\begin{tabular}{@{\hskip 1mm}c@{\hskip 1mm}c}
{\rotatebox{90}{\hspace{3mm}Input}}
\includegraphics[width=0.49\columnwidth]{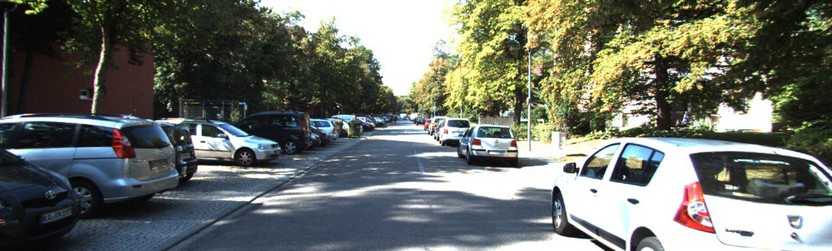} &
\includegraphics[width=0.49\columnwidth]{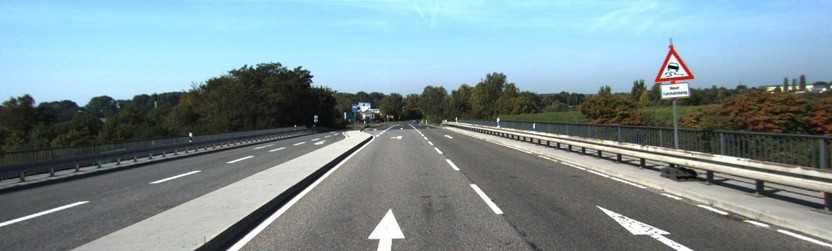} \\

{\rotatebox{90}{\hspace{0mm}\scriptsize MD2 (\textbf{M})}}
\includegraphics[width=0.49\columnwidth]{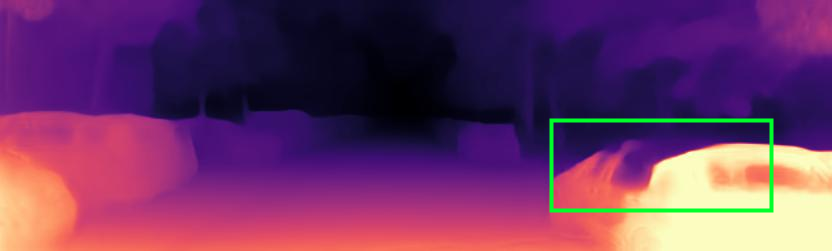} &
\includegraphics[width=0.49\columnwidth]{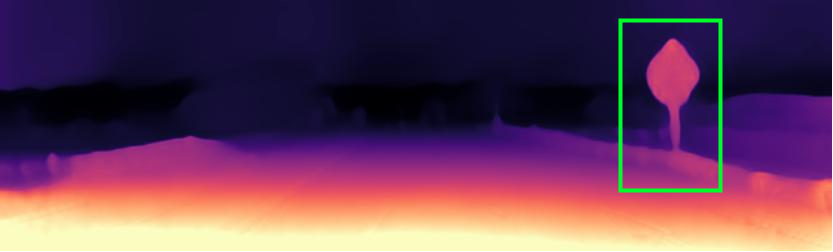} \\

{\rotatebox{90}{\hspace{1mm}\scriptsize Ours (\textbf{M}) }}
\includegraphics[width=0.49\columnwidth]{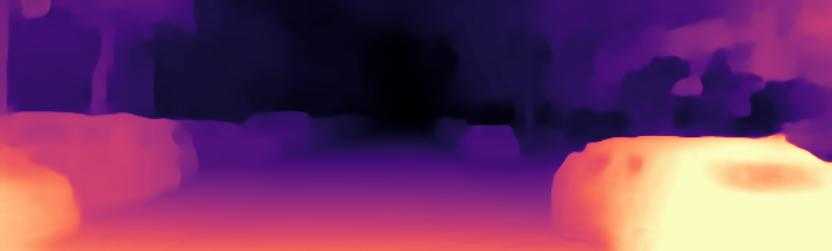} &
\includegraphics[width=0.49\columnwidth]{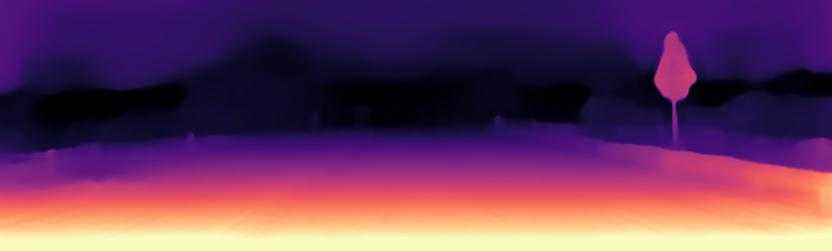} \\

\end{tabular}



	\caption{\textbf{Monodepth2 Failure cases.} Although trained on the same loss function as the monocular trained ({\bf M}) Monodepth2 \cite{monodepth2}, our method succeeds in estimating depth for the reflective car roof \emph{(Left)} and the difficult to delineate street sign \emph{(Right)}.}
	\label{fig:md2_failure}
\end{figure}

\vspace{-3pt}
The results for the  experiment are presented in Table~\ref{tab:kitti_eigen}. When comparing our method (grayed row in Table~\ref{tab:kitti_eigen}) on the KITTI 2015 data set~\cite{Geiger2012CVPR} (using Eigen~\cite{eigen2015predicting} split), we observe that we outperform all existing self-supervised monocular trained methods by a significant margin.   Compared to other methods that rely on stronger supervision signals (e.g., stereo supervision and mono+stereo supervision), our approach is competitive, producing comparable results to the current state of the art method Monodepth2. As can be seen in Figure \ref{fig:qual-results} our method shows sharper results on thinner structures such as poles than the baseline Monodepth2. In general, Monodepth2 (Mono and Mono+Stereo) struggles with thin structures that overlap with foliage, while our method is able to accurately estimate the depth of these smaller details. We attribute this to the combination of the dilated convolutions and the contextual information from the self-attention module. As can be seen in car windows, Monodepth2 and our method struggle to predict the depth on glassy reflective surfaces. However, this is a common issue observed in self-supervised methods because they cannot accurately predict depth for transparent surfaces since the photometric reprojection/warping error is ill-defined for such materials/surfaces. For instance, in the example of car windows, the correct depth that would minimise the photometric reprojection loss is actually the depth from the car interior, instead of the glass depth, as would be recorded by the ground truth LiDAR. When comparing our method against some specific error cases for Monodepth2~\cite{monodepth2} (Figure~\ref{fig:md2_failure}), we can see that our method succeeds in estimating depth of the highly reflective car roof (\emph{left}) and successfully disentangles the street sign from the background (\emph{right}). This can be explained by the extra context and receptive field afforded by the self-attention context module as well as the regularisation provided by the discrete disparity volume.

\subsection{Make3D Results}

Table~\ref{tab:make3d} presents the quantitative results for the Make3D data set~\cite{saxena2009make3d} using our model trained on KITTI2015. We follow the same testing protocol as Monodepth2~\cite{monodepth2} and methods are compared using the evaluation criteria outline in~\cite{godard2017unsupervised}. It can be seen in Table~\ref{tab:make3d} that our method produces superior results compared with previous methods that also rely on self-supervision.

\subsection{Ablation Study}\label{sec:ablation}


\begin{table*}[t!]
	\centering
	\resizebox{0.9\textwidth}{!}{
		\begin{tabular}{|l|c|c|c|c||c|c|c|c|c|c|}
			\hline
			Backbone & Self-Attn  & DDV & \cellcolor{col1}Abs Rel & \cellcolor{col1}Sq Rel & \cellcolor{col1}RMSE & \cellcolor{col1}RMSE log & \cellcolor{col2}$\delta < 1.25 $ & \cellcolor{col2}$\delta < 1.25^{2}$ & \cellcolor{col2}$\delta < 1.25^{3}$ \\ \hline
			Baseline (MD2 ResNet18) & \xmark  & \xmark & 0.115 & 0.903 & 4.863 & 0.193 & 0.877  & 0.959  & 0.981 \\
			ResNet18  & \xmark & \checkmark & 0.112  & 0.838  & 4.795 & 0.191 & 0.877 & 0.960 & 0.981 \\
			ResNet18  & \checkmark  & \xmark & 0.112 &	0.845 &	4.769 &	0.19 &	0.877 &	0.96 &	{\bf 0.982} \\
			ResNet18  & \checkmark  & \checkmark & 0.111  &   0.941  & 4.817 & 0.189 &  0.885 &  0.961 & 0.981  \\
			ResNet101 w/ Dilated Conv & \xmark & \xmark & 0.110 & 0.876 & 4.853 & 0.189 & 0.879 & 0.961  & {\bf 0.982} \\
			ResNet101 w/ Dilated Conv & \xmark & \checkmark & 0.110 & 0.840 & 4.765 & 0.189 & 0.882 & 0.961 & {\bf 0.982}\\
			ResNet101 w/ Dilated Conv & \checkmark & \xmark & 0.108 & {\bf 0.808}& 4.754 & {\bf 0.185} & 0.885 & {\bf 0.962} & {\bf 0.982} \\
			ResNet101 w/ Dilated Conv & \checkmark & \checkmark & {\bf 0.106} & 0.861 & {\bf 4.699} & {\bf 0.185} & {\bf 0.889} & {\bf 0.962} & {\bf 0.982} \\													
			\hline
		\end{tabular}
	}
	\caption{\textbf{Ablation Study.} Results for different versions of our model with comparison to our baseline model Monodepth2 \cite{monodepth2}(MD2 ResNet18). We evaluate the impact of the Discrete Disparity Volume (DDV), Self-Attention Context module and the larger network architecture. All models were trained with Monocular self-supervision. Metrics indicated by red: \emph{lower is better}, Metrics indicated by blue: \emph{higher is better}}
	\label{tab:ablation}
	
\end{table*}

Table~\ref{tab:ablation} shows an ablation study of our method, where we start from the baseline Monodepth2~\cite{monodepth2} (row 1).  Then, by first adding DDV (row 2) and both self attention and DDV (row 3), we observe a steady improvement in almost all evaluation measures. We then switch the underlying encoding model ResNet-18 to ResNet-101 with dilated convolutions in row 4. Rows 5 and 6 show the addition of DDV and then both self-attention and DDV, respectively, again with a steady improvement of evaluation results in almost all evaluation measures.  The DDV on the smaller ResNet-18 model provides a large improvement over the baseline in the \emph{absolute relative} and \emph{squared relative} measures. However, ResNet-101 shows only a small improvement over the baseline when using the DDV. The Self-Attention mechanism drastically improves the close range accuracy ($\delta < 1.25$) for both backbone models. The significantly larger improvement of the self-attention module in the ResNet-101 model (row 6), is likely because of the large receptive field produced by the dilated convolutions, which increases the amount of contextual information that can be computed by the self-attention operation.
 


\begin{table}
    \centering
    \resizebox{\linewidth}{!}{
      \begin{tabular}{|l|c||c|c|c|c|}
      \hline
       & Type & \cellcolor{col1}Abs Rel & \cellcolor{col1}Sq Rel  & \cellcolor{col1}RMSE & \cellcolor{col1}$\text{log}_{10}$ \\
      \hline
      Karsch \cite{karsch2014depth} & D & 0.428 & 5.079 & 8.389 & 0.149 \\
      Liu \cite{liu2014discrete}& D & 0.475 & 6.562 & 10.05 & 0.165 \\
      Laina \cite{laina2016deeper}& D & {\bf 0.204} & {\bf 1.840} & {\bf 5.683} & {\bf 0.084} \\ \hline
      Monodepth \cite{godard2017unsupervised} & S & 0.544 & 10.94 & 11.760 & 0.193 \\
      Zhou \cite{zhou2017unsupervised} & M &  0.383 & 5.321 & 10.470 & 0.478 \\
      DDVO \cite{wang2017learning} & M & 0.387 & 4.720 & 8.090 & 0.204 \\
      Monodepth2 \cite{monodepth2} & M & 0.322 & 3.589 & 7.417 & 0.163  \\
      \textbf{Ours} & M &   \textbf{0.297} &  \textbf{2.902} &  \textbf{7.013} &  \textbf{0.158} \\ %
    
      \hline
      \end{tabular}
    }
    \vspace{2.0pt}
    \caption{{\bf Make3D results.} All self-supervised mono (M) models use median scaling.}
    \label{tab:make3d}
    \vspace{-8pt}
\end{table}

\subsection{Self-attention and Depth Uncertainty}

\begin{figure}
	\centering
	\newcommand{\imlabel}[2]{\includegraphics[width=0.49\columnwidth]{#1}%
\raisebox{2pt}{\makebox[-2pt][r]{\footnotesize #2}}}
\newcommand{\turnheightnew}{0.195\columnwidth}
\centering

\begin{tabular}{@{\hskip 1mm}c@{\hskip 1mm}c}
\includegraphics[width=0.49\columnwidth]{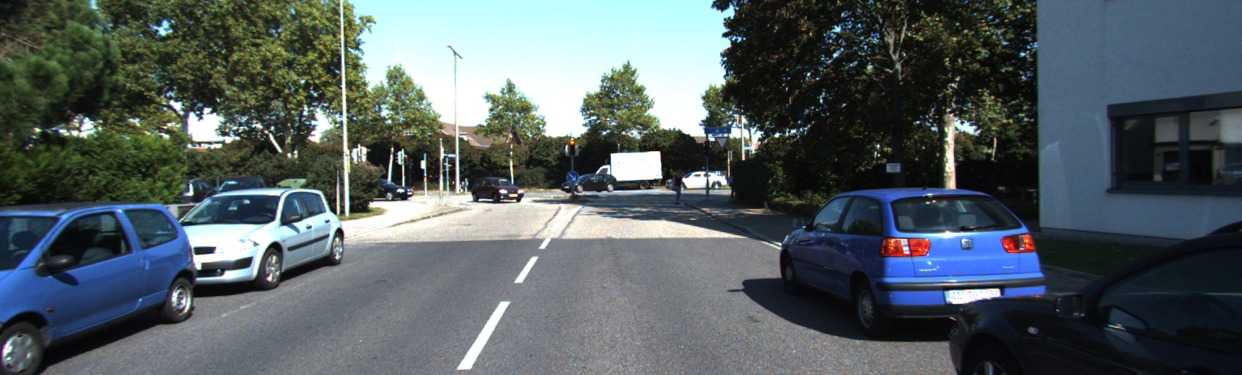} &
\includegraphics[width=0.49\columnwidth]{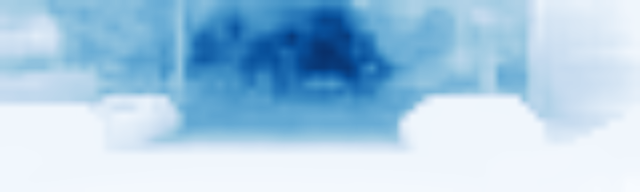} \\

{
}
\includegraphics[width=0.49\columnwidth]{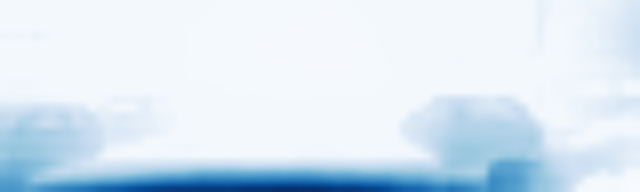} &
\includegraphics[width=0.49\columnwidth]{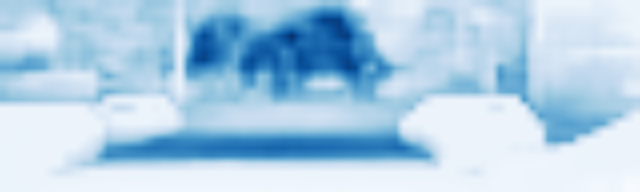} \\


\includegraphics[width=0.49\columnwidth]{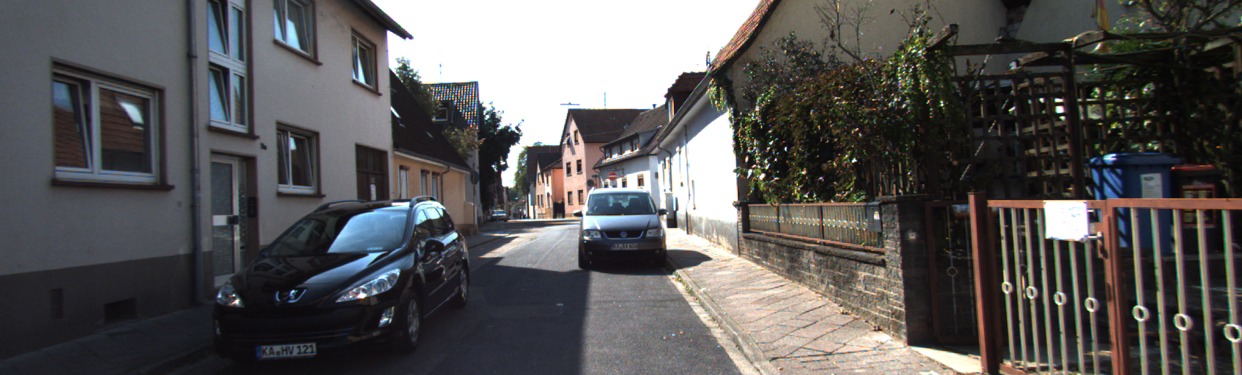} &
\includegraphics[width=0.49\columnwidth]{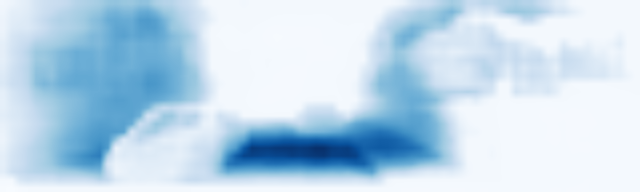} \\

{
}
\includegraphics[width=0.49\columnwidth]{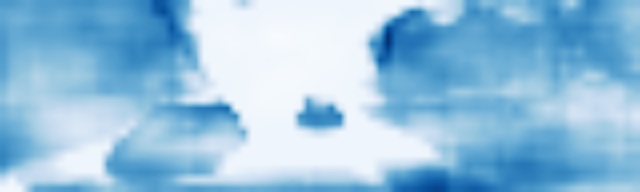} &
\includegraphics[width=0.49\columnwidth]{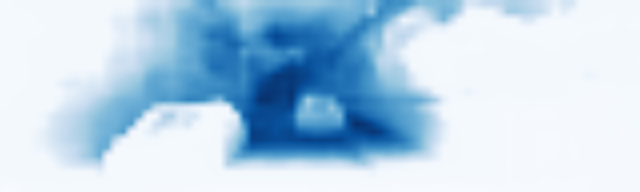} \\


\end{tabular}
	\caption{\textbf{Attention maps from our network}. Subset of the attention maps produced by our method. Blue indicates region of attention.}
	\label{fig:attn_maps}
\end{figure}

\begin{figure}
	\centering
	\newcommand{\imlabel}[2]{\includegraphics[width=0.49\columnwidth]{#1}%
\raisebox{2pt}{\makebox[-2pt][r]{\footnotesize #2}}}
\newcommand{\turnheightnew}{0.195\columnwidth}
\centering

\begin{tabular}{@{\hskip 1mm}c@{\hskip 1mm}c}

\includegraphics[width=0.49\columnwidth]{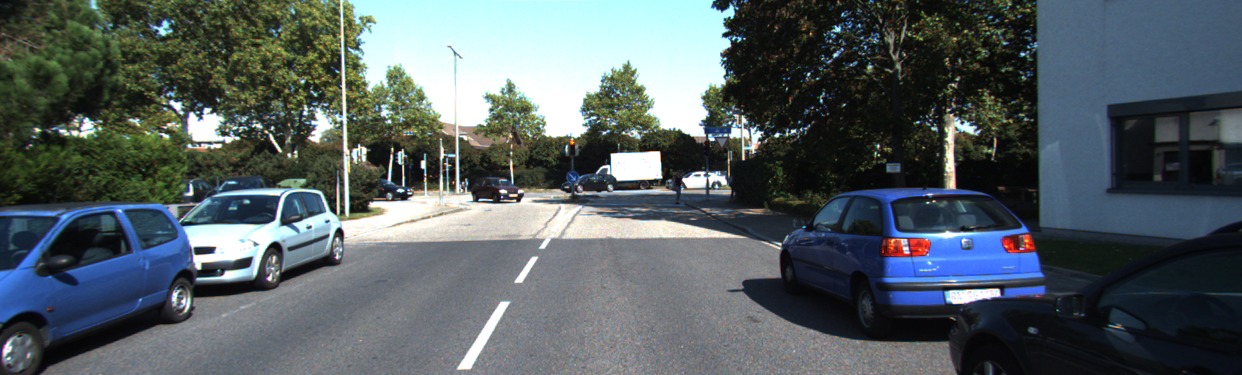} &
\includegraphics[width=0.49\columnwidth]{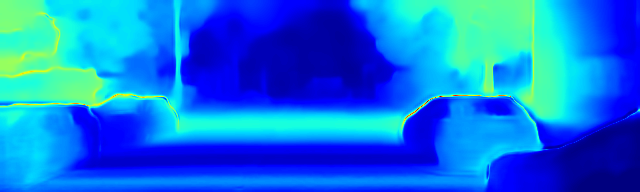} \\

\includegraphics[width=0.49\columnwidth]{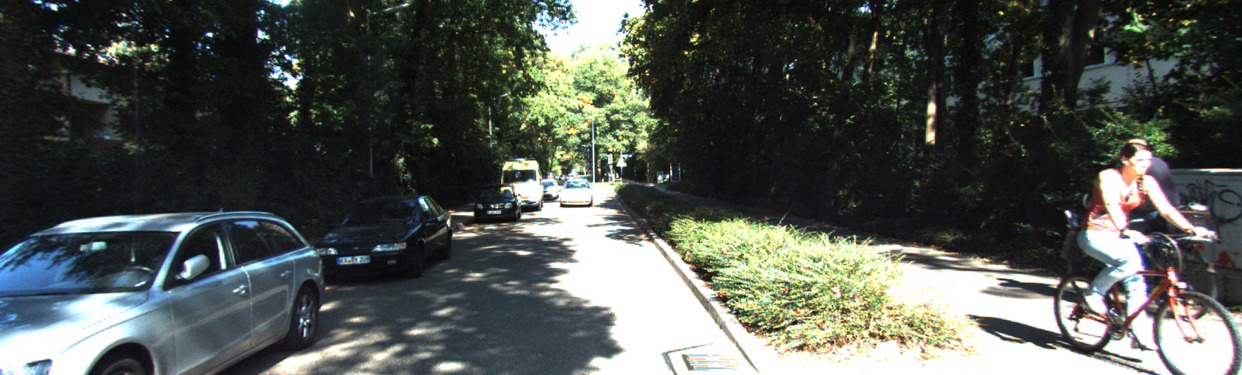} &
\includegraphics[width=0.49\columnwidth]{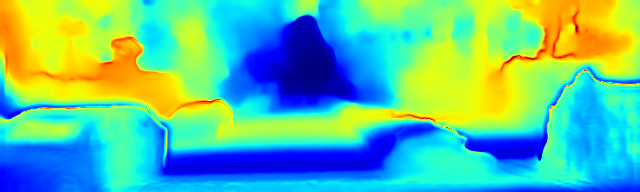} \\

\includegraphics[width=0.49\columnwidth]{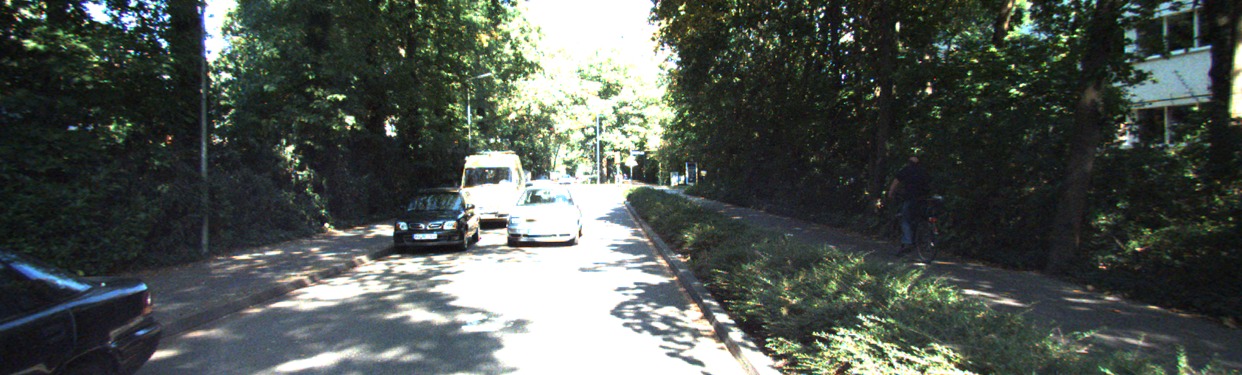} &
\includegraphics[width=0.49\columnwidth]{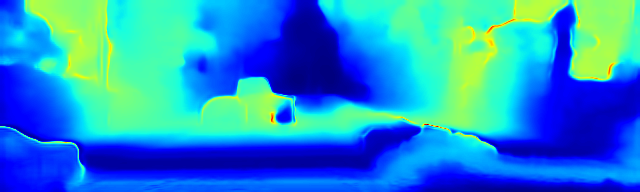} \\

\includegraphics[width=0.49\columnwidth]{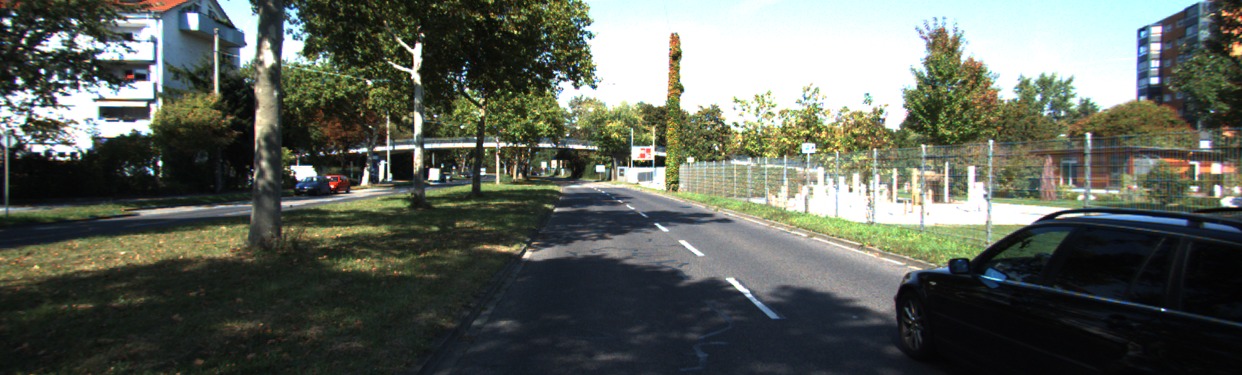} &
\includegraphics[width=0.49\columnwidth]{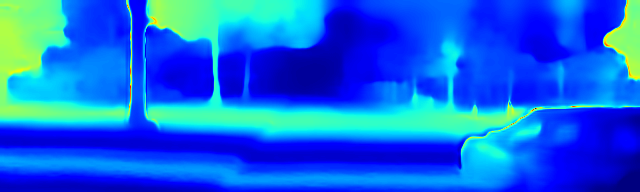} \\

\includegraphics[width=0.49\columnwidth]{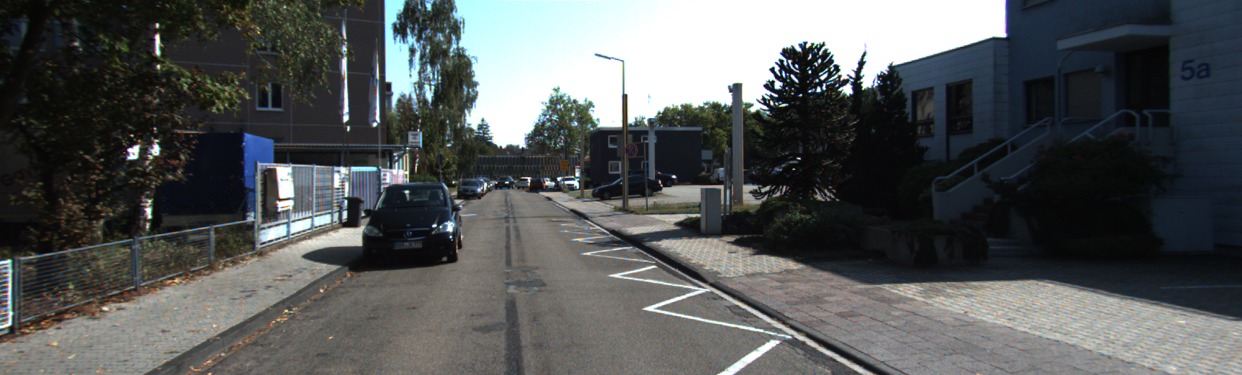} &
\includegraphics[width=0.49\columnwidth]{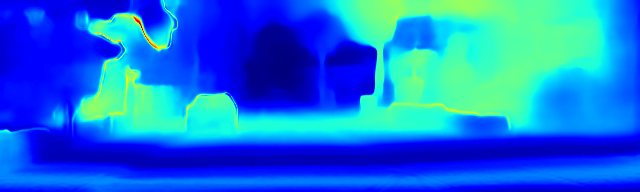} \\

\end{tabular}
	\caption{\textbf{Uncertainty from our network}. The Discrete Disparity Volume allows us to compute pixel-wise depth uncertainty. Blue indicates areas of low uncertainty, green/red regions indicate areas of high/highest uncertainty.}
     \vspace*{-5mm}
	\label{fig:uncert}
\end{figure}

While the self-attention module and DDV together provide significant quantitative and qualitative improvements, they also provide secondary functions. The attention maps (Eq.~\ref{eq:self_attention}) from the self-attention module can be visualized to interrogate the relationships between objects and disparity learnt by the model. The attention maps highlight non-contiguous image regions (Fig.~\ref{fig:attn_maps}), focusing on either foreground, midground or background regions. The maps also tend to highlight either distant objects or stationary visual objects, like cars. Moreover, as the DDV encodes a probability over a disparity ray, using discretized bins, it is possible to compute the uncertainty for each ray by measuring the variance of the probability distribution. Figure~\ref{fig:uncert} shows a trend where uncertainty increases with distance, up until the background image regions, which are estimated as near-infinite to infinite depth with very low uncertainty. This has also been observed in supervised models that are capable of estimating uncertainty \cite{liu2019neural}. Areas of high foliage and high shadow (row 2) show very high uncertainty, likely attributed to the low contrast and lack of textural detail in these regions.




\section{Conclusion}
\vspace{-1pt}

 In this paper we have presented a method to address the challenge of learning to predict accurate disparities solely from monocular video. By incorporating a self-attention mechanism to improve the contextual information available to the model, we have achieved state of the art results for monocular trained self-supervised depth estimation on the KITTI 2015~\cite{Geiger2012CVPR} dataset. Additionally,  we regularised the training of the model by using a discrete disparity volume, which allows us to produce more robust and sharper depth estimates and to compute pixel-wise depth uncertainties. In the future, we plan to investigate the benefits of incorporating self-attention in the pose model as well as using the estimated uncertainties for outlier filtering and volumetric fusion.

\section{Acknowledgment}
This research was in part supported by the Data to Decisions Cooperative Research Centre (A.J) and the Australian Research Council through grants DP180103232, CE140100016.  G.C. acknowledges the support by the Alexander von Humboldt-Stiftung for the renewed research stay sponsorship.
\vspace{-2pt}

{\small
	\bibliographystyle{ieee_fullname}
	\bibliography{egbib}
}

\clearpage

\begin{appendices}

\section{KITTI Improved Ground Truth}

The evaluation method that was introduced by Eigen \etal~\cite{eigen2015predicting} uses reprojected LiDAR points to create the ground truth depth images. However, the reprojections do not handle occlusions, non-rigid motion or motion from the camera. Uhrig \etal~\cite{uhrig2017sparse} introduced an improved set of high quality ground truth depth maps for the KITTI dataset. These high quality images are instead reprojected using 5 consecutive LiDAR frames and uses the stereo images for better handling of occlusions. To obviate the need of retraining, as with other work \cite{monodepth2}, we use a modified Eigen~\cite{eigen2015predicting} test split on the images that overlap between these datasets. This results in 652 (93\%) of the 697 original test frames being retained. We use the same evaluation strategy and metrics as discussed in the Experiments section of the main paper. The results of this analysis can be found in Table \ref{tab:kitti_improved_gt}. 

\section{Network Architecture}
For all experiments, except where noted, we use a ResNet-101 encoder model with pretrained ImageNet weights. This model has been modified to use atrous/dilation convolutions ~\cite{chen2017rethinking} in the final two residual blocks. We use rectified linear activation (ReLU) in the encoding model and the Exponential Linear Unit (ELU) in the decoder. Skip connections are applied to the two intermediate outputs between the encoder and decoder. As the internal resolution is much larger than that of the ResNet-18 used by Monodepth2 \cite{monodepth2} ($\frac{1}{8}$ scale compared with $\frac{1}{32}$ scale), a skip connection is not required for the smallest output resolution. For the pose model, we use the same ResNet-18 and pose decoder defined by Monodepth2 \cite{monodepth2}. The full depth network architecture can be found in Table \ref{tab:network}.

\begin{table}[!htbp]
  \centering
  \resizebox{1.0\columnwidth}{!}{
  \begin{tabular}[t]{l}

\begin{tabular}[t]{|l|l|l|l|l|l|l|l|}
\hline
\multicolumn{8}{|l|}{\textbf{Depth Network} (Number of Parameters: 51.34M)} \\
\hline
\textbf{layer} & \textbf{k} & \textbf{s} & \textbf{ch} & \textbf{dilation} & \textbf{res} & \textbf{input}   & \textbf{activation}    \\ \hline

conv1       & 3      & 1      & 64    & 2    & 1    & image                         & ReLU \\
conv2       & 3      & 1      & 64    & 1    & 2    & conv1                         & ReLU  \\
conv3       & 3      & 1      & 128   & 1    & 2    & conv2                         & ReLU  \\
maxpool     & 3      & 2      & 128   & 1    & 2    & conv2                         & ReLU  \\

\hline

res1        & 3      & 1      & 256    & 1    & 4    & conv3                         & ReLU  \\
res2        & 3      & 2      & 512   & 1     & 8    & res1                          & ReLU  \\
res3        & 3      & 1      & 1024   & 2    & 8    & res2                          & ReLU  \\
res4        & 3      & 1      & 2048   & 4    & 8    & res4                          & ReLU  \\

\hline
context     & 3      & 1      & 512   & 1    & 8    & res4                          & Self-Attn  \\
ddv4        & 3      & 1      & 128   & 1    & 8    & context                       & Linear  \\
disp4       & 3      & 1      & 1     & 1    & 8    & ddv1                          & softmax  \\

\hline

upconv3       & 3      & 1      & 64    & 1    & 8  & ddv4                          & ELU  \\
deconv3       & 3      & 1      & 64    & 1    & 4  & upconv3$\uparrow$, res1       & ELU \\ 
ddv3          & 3      & 1      & 128   & 1    & 4  & deconv3                       & Linear  \\
disp3         & 3      & 1      & 1     & 1    & 4  & ddv3                          & softmax  \\

\hline

upconv2       & 3      & 1      & 64    & 1    & 4  & deconv3                       & ELU  \\
deconv2       & 3      & 1      & 64    & 1    & 2  & upconv2$\uparrow$, conv3      & ELU \\ 
ddv2          & 3      & 1      & 128   & 1    & 2  & deconv2                       & Linear  \\
disp2         & 3      & 1      & 1     & 1    & 2  & ddv2                          & softmax  \\
\hline

upconv1       & 3      & 1      & 32    & 1    & 2  & deconv2                       & ELU  \\
deconv1       & 3      & 1      & 32    & 1    & 1  & upconv1$\uparrow$             & ELU \\ 
ddv1          & 3      & 1      & 128   & 1    & 1  & deconv1                       & Linear  \\
disp1         & 3      & 1      & 1     & 1    & 1  & ddv1                          & softmax  \\
\hline

\end{tabular}  \\


\end{tabular} 
 }
  \vspace{2pt}
  \caption{\textbf{Network architecture}. This table details the kernel size (\textbf{k}), stride (\textbf{s}), output channels (\textbf{ch}) dilation factor (\textbf{dilation}), resolution scale (\textbf{res}), input features for each layer (\textbf{input}) and activation function (\textbf{activation}) used in our model. Layers marked with $\uparrow$ represent a $2\times$ nearest-neighbour upsampling before passing to the convolutional layer. Residual blocks are denoted by $res*$ naming convention. Each convolution and residual block also uses batch normalisation in the form of a inplace activated batch normalisation~\cite{rota2018place}. The self-attention module (\emph{context}) is denoted as having an activation of \emph{Self-Attn}.
  }
  \vspace{40pt}
\label{tab:network}
\end{table}

\section{Additional Qualitative Results}

\begin{figure*}[!htbp]
    \centering
    \resizebox{\textwidth}{!}{
    	\newcommand{\turnheightnew}{0.195\columnwidth}

\centering

\begin{tabular}{@{\hskip 2mm}c@{\hskip 2mm}c@{\hskip 2mm}c@{\hskip 2mm}c@{\hskip 2mm}c@{}}

{\rotatebox{90}{\hspace{4mm}Input}} &
\includegraphics[height=\turnheightnew]{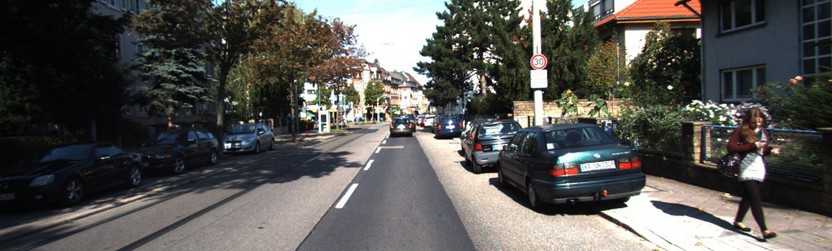} &
\includegraphics[height=\turnheightnew]{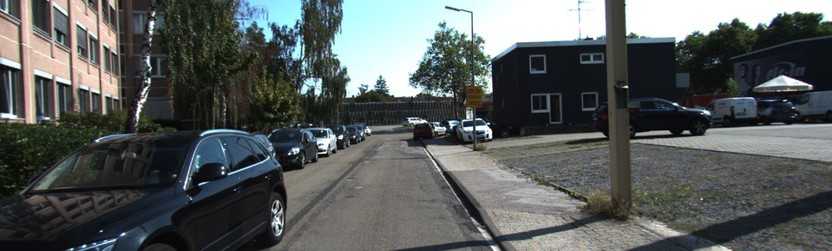} &
\includegraphics[height=\turnheightnew]{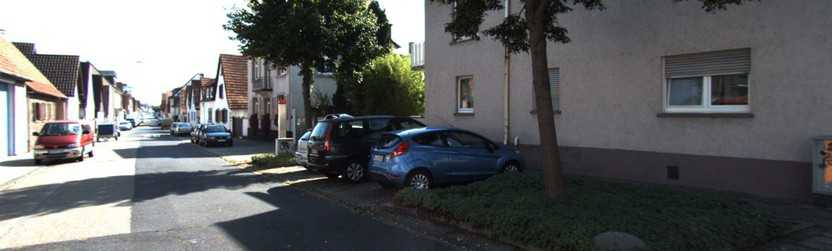} &
\includegraphics[height=\turnheightnew]{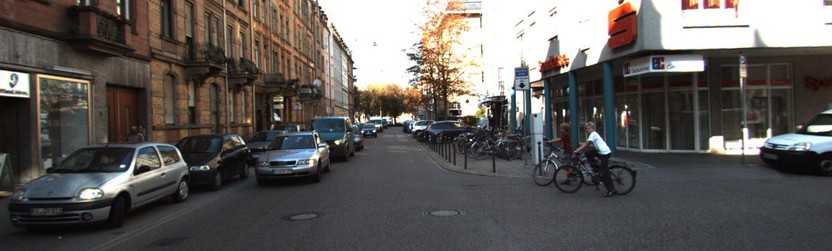}\\

{\rotatebox{90}{\hspace{0mm}\scriptsize
{Monodepth \cite{godard2017unsupervised}}}} &
\includegraphics[height=\turnheightnew]{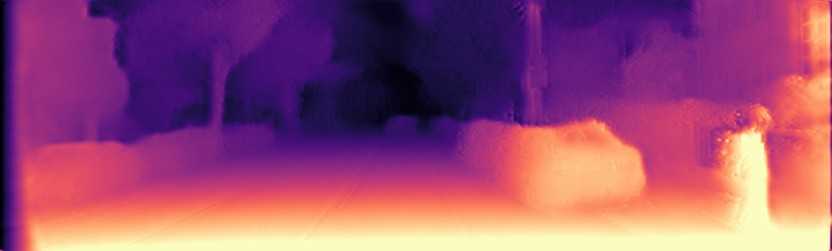} &
\includegraphics[height=\turnheightnew]{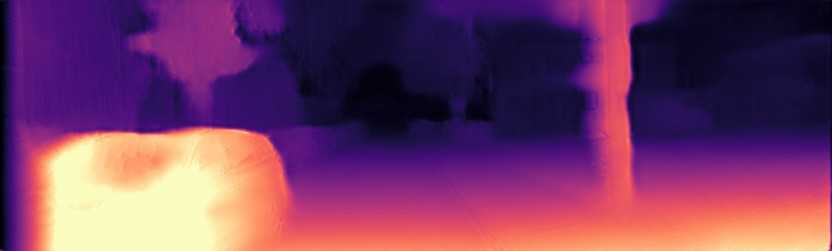} &
\includegraphics[height=\turnheightnew]{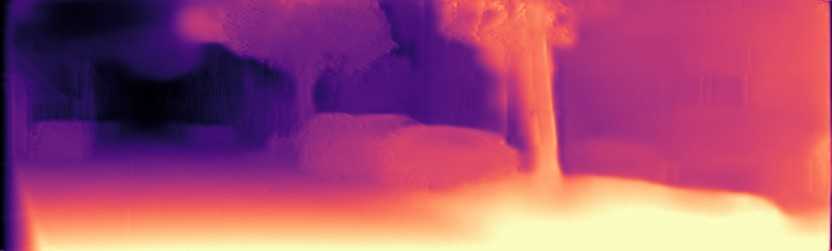} &
\includegraphics[height=\turnheightnew]{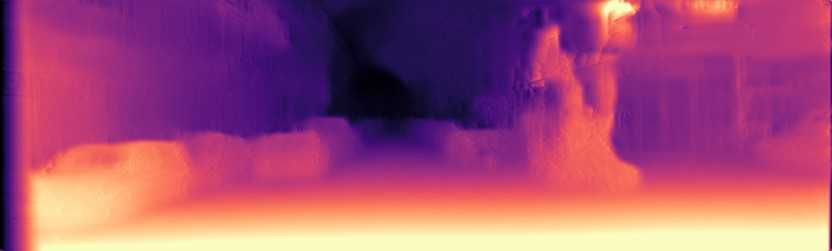} \\

{\rotatebox{90}{\hspace{0mm}\scriptsize
{Zhou \ea~\cite{zhou2017unsupervised}}}} &
\includegraphics[height=\turnheightnew]{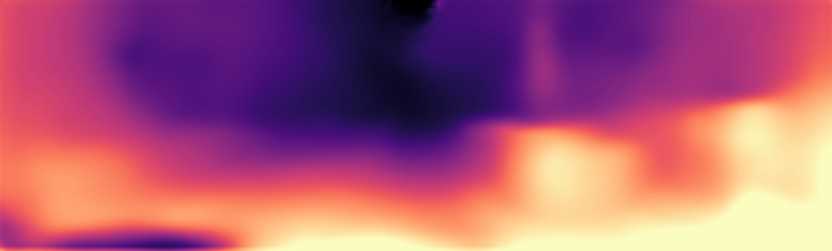} &
\includegraphics[height=\turnheightnew]{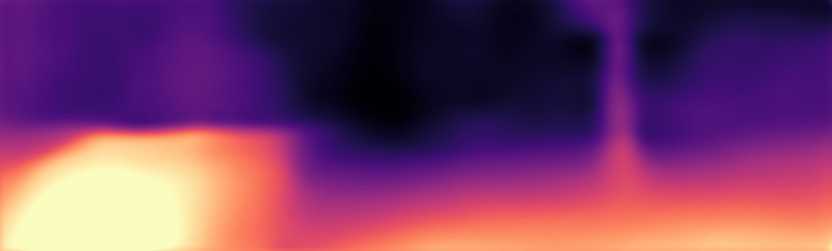} &
\includegraphics[height=\turnheightnew]{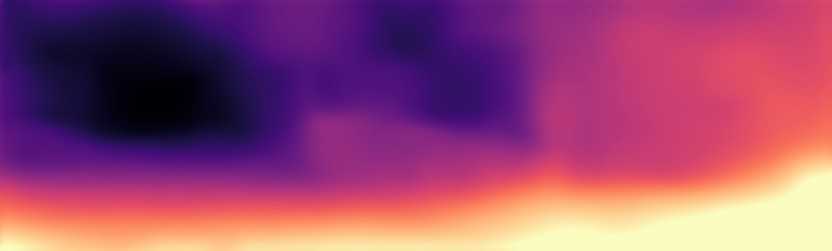} &
\includegraphics[height=\turnheightnew]{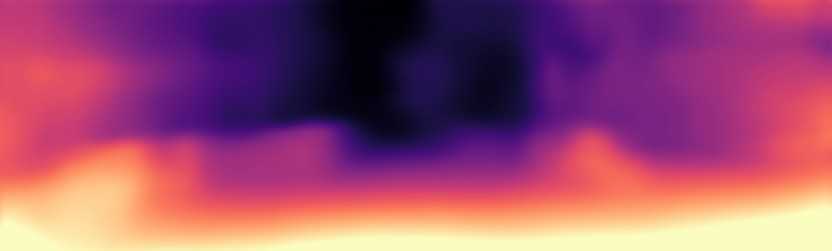} \\

{\rotatebox{90}{\hspace{2mm}\scriptsize
{DDVO~\cite{wang2017learning}}}} &
\includegraphics[height=\turnheightnew]{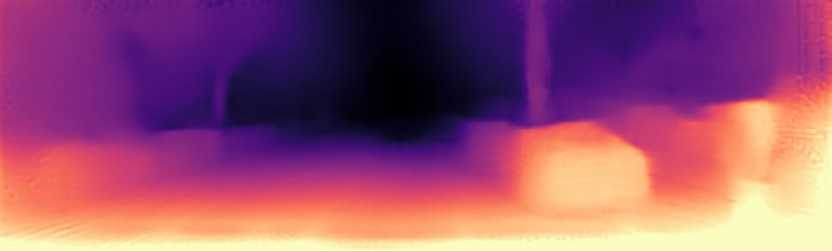} &
\includegraphics[height=\turnheightnew]{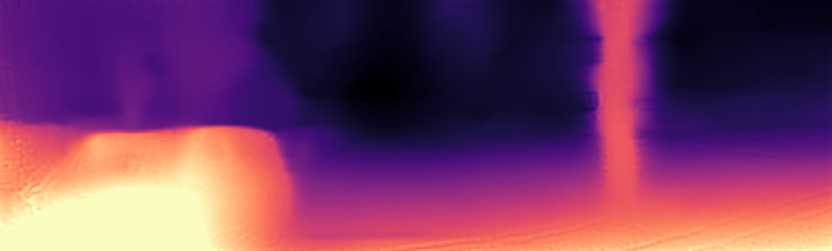} &
\includegraphics[height=\turnheightnew]{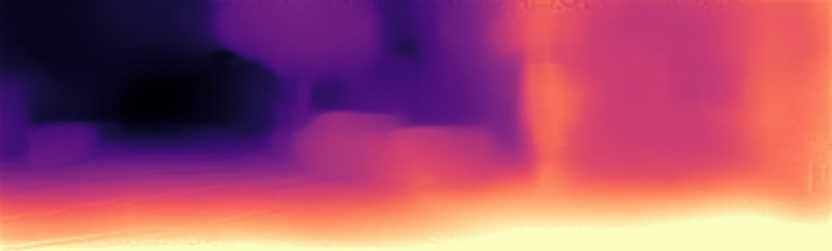} &
\includegraphics[height=\turnheightnew]{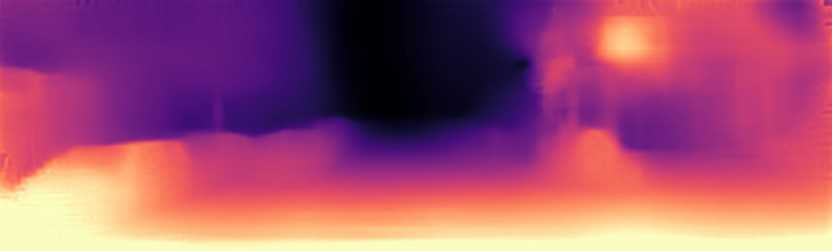} \\

{\rotatebox{90}{\hspace{2mm}\scriptsize
{GeoNet~\cite{geonet2018}}}} &
\includegraphics[height=\turnheightnew]{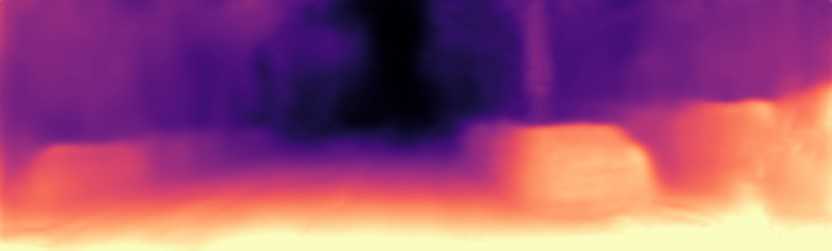} &
\includegraphics[height=\turnheightnew]{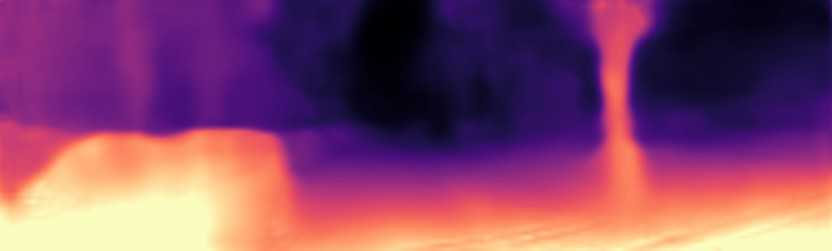} &
\includegraphics[height=\turnheightnew]{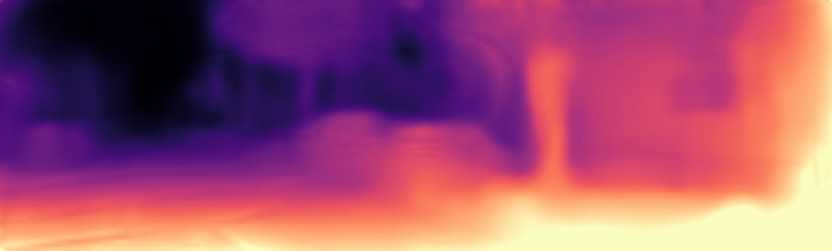} &
\includegraphics[height=\turnheightnew]{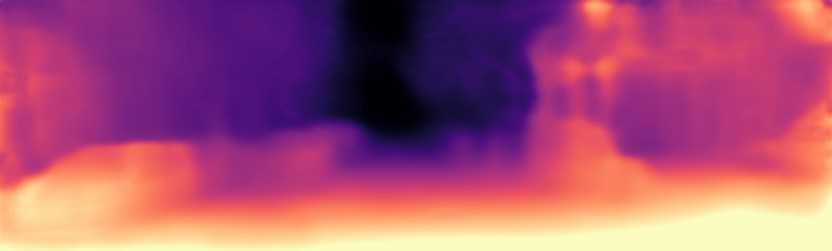} \\

{\rotatebox{90}{\hspace{0mm}\scriptsize {Zhan \ea~\cite{zhanst2018}}}} &
\includegraphics[height=\turnheightnew]{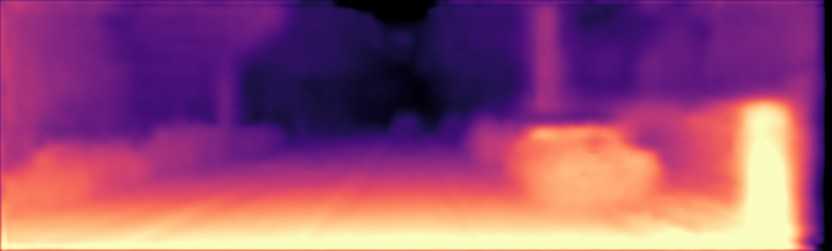} &
\includegraphics[height=\turnheightnew]{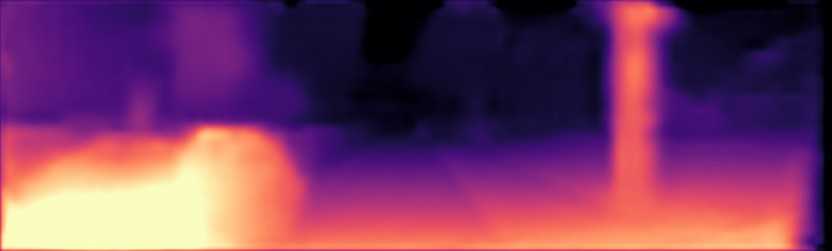} &
\includegraphics[height=\turnheightnew]{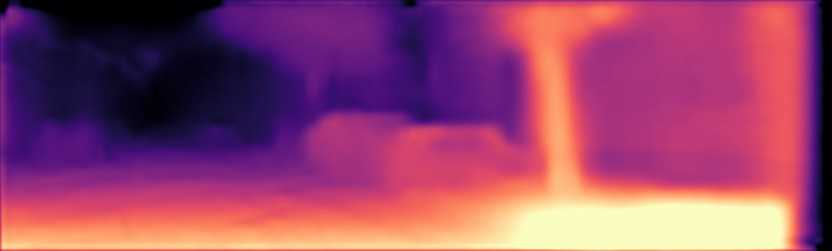} &
\includegraphics[height=\turnheightnew]{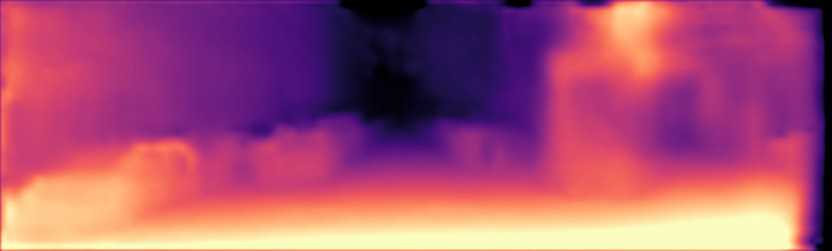} \\

{\rotatebox{90}{\hspace{-1mm}\scriptsize Ranjan \ea~\cite{ranjan2018adversarial}}} &
\includegraphics[height=\turnheightnew]{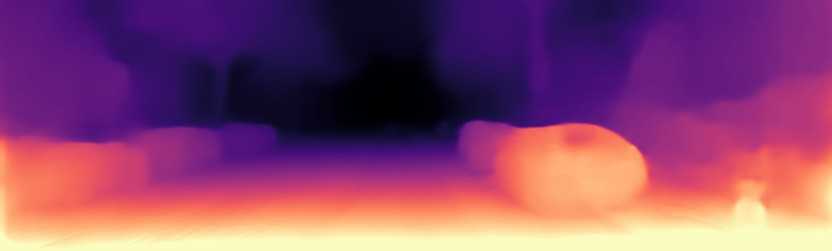} &
\includegraphics[height=\turnheightnew]{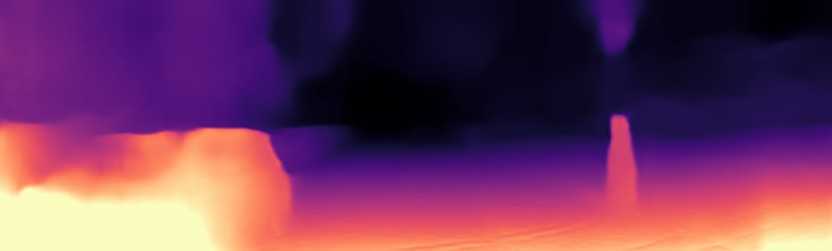} &
\includegraphics[height=\turnheightnew]{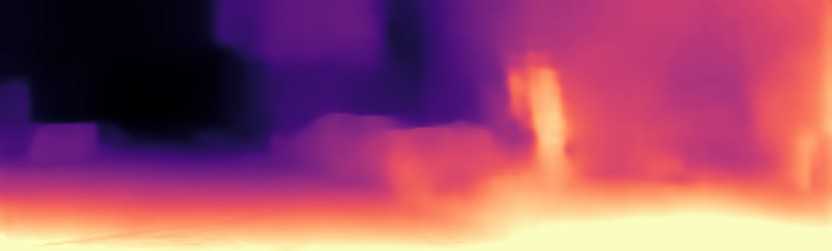} &
\includegraphics[height=\turnheightnew]{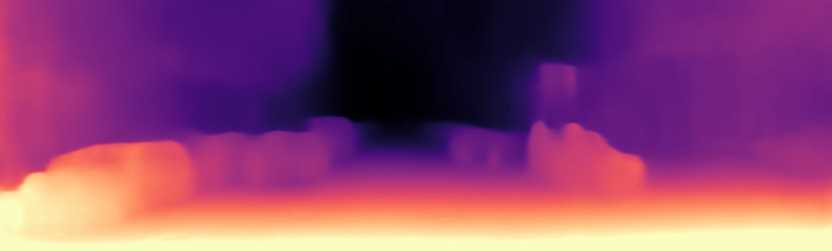} \\

{\rotatebox{90}{\hspace{0mm} \scriptsize 3Net - R50 \cite{luo2018every}}} &
\includegraphics[height=\turnheightnew]{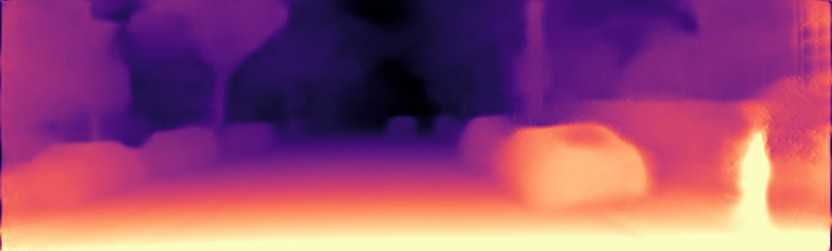} &
\includegraphics[height=\turnheightnew]{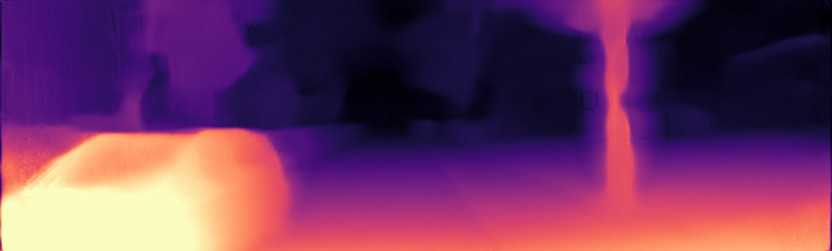} &
\includegraphics[height=\turnheightnew]{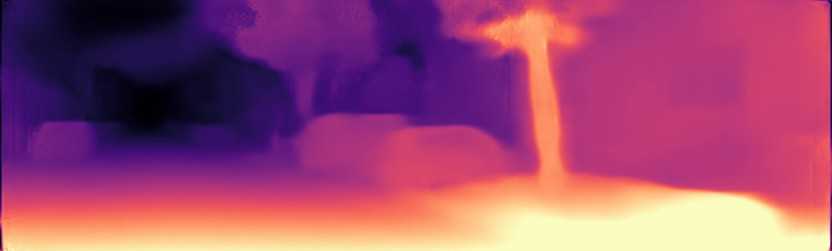} &
\includegraphics[height=\turnheightnew]{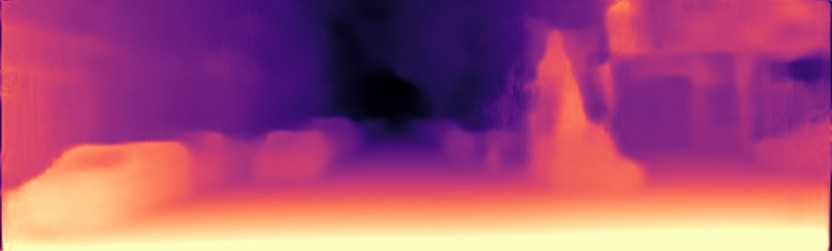} \\ 

{\rotatebox{90}{\hspace{-3mm} \scriptsize EPC++ (MS) \newline \cite{luo2018every}}} &
\includegraphics[height=\turnheightnew]{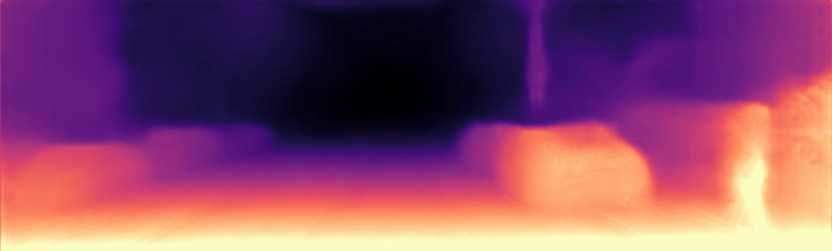} &
\includegraphics[height=\turnheightnew]{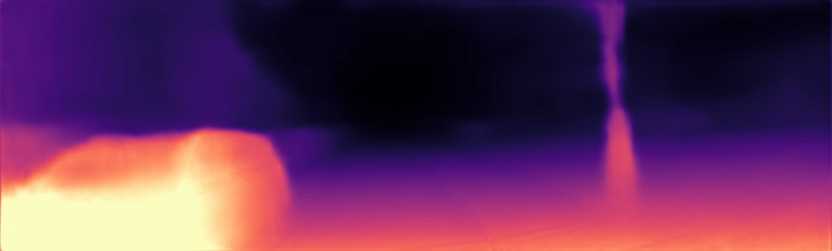} &
\includegraphics[height=\turnheightnew]{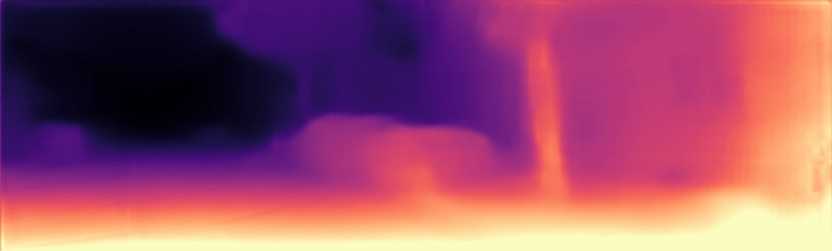} &
\includegraphics[height=\turnheightnew]{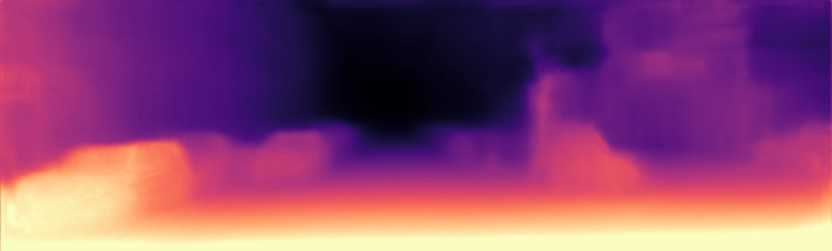} \\ 

{\rotatebox{90}{\scriptsize\hspace{3mm}\textbf{MD2 M}}} &
\includegraphics[height=\turnheightnew]{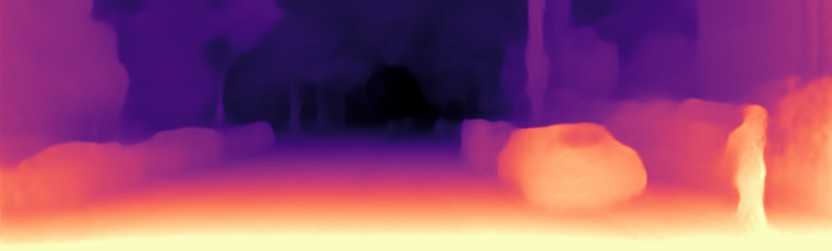} &
\includegraphics[height=\turnheightnew]{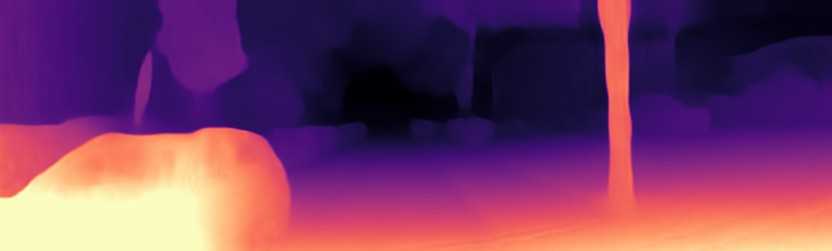} &
\includegraphics[height=\turnheightnew]{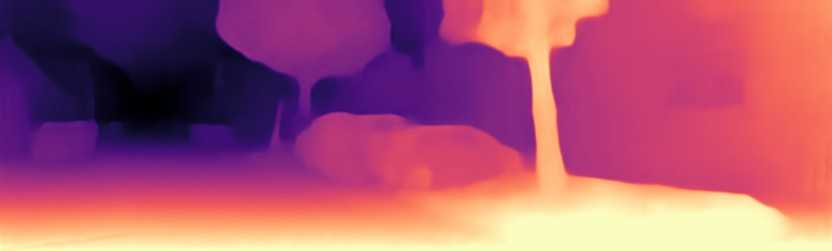} &
\includegraphics[height=\turnheightnew]{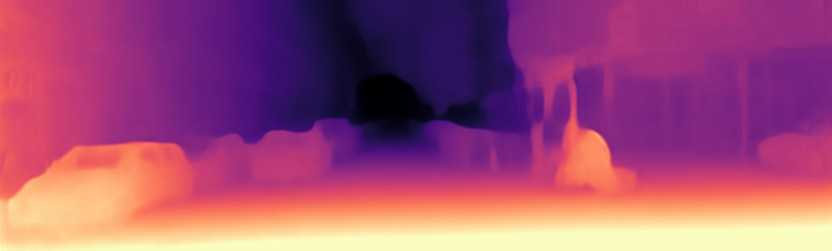} \\


{\rotatebox{90}{\scriptsize\hspace{3mm}\textbf{MD2 S}}} &
\includegraphics[height=\turnheightnew]{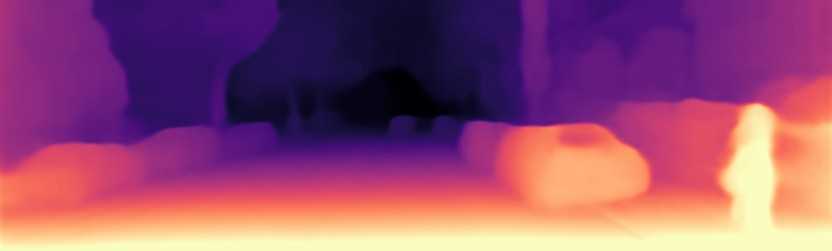} &
\includegraphics[height=\turnheightnew]{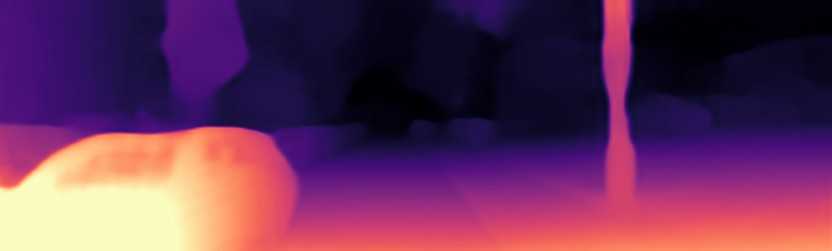} &
\includegraphics[height=\turnheightnew]{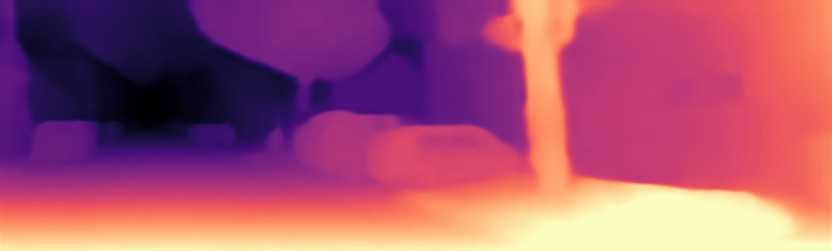} &
\includegraphics[height=\turnheightnew]{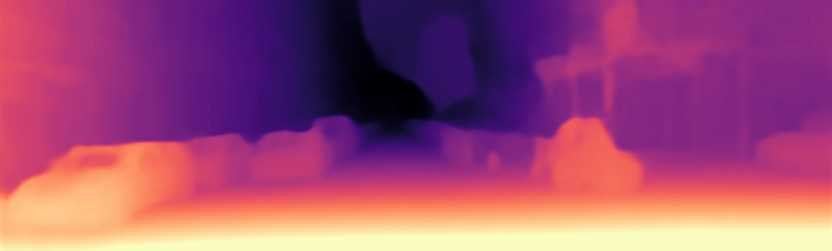} \\

{\rotatebox{90}{\scriptsize\hspace{3mm}\textbf{MD2 MS}}} &
\includegraphics[height=\turnheightnew]{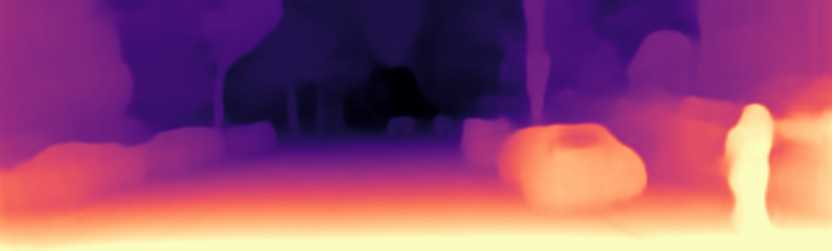} &
\includegraphics[height=\turnheightnew]{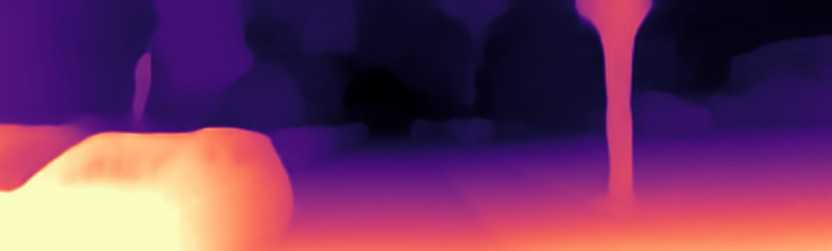} &
\includegraphics[height=\turnheightnew]{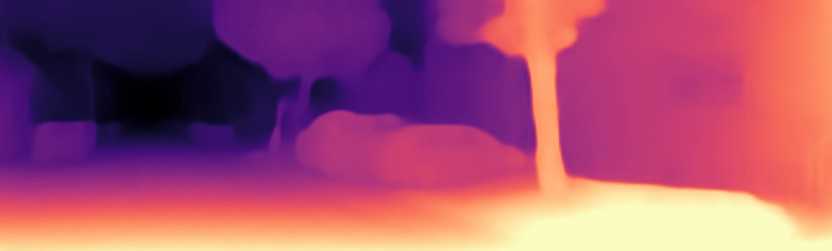} &
\includegraphics[height=\turnheightnew]{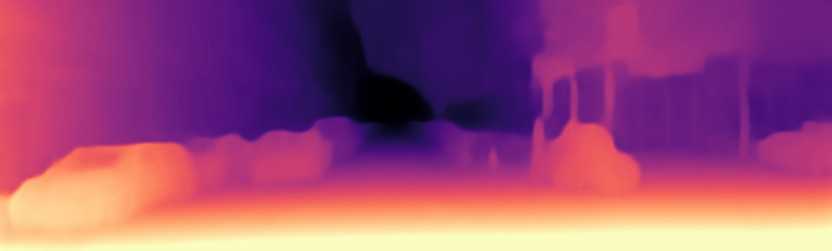} \\

{\rotatebox{90}{\scriptsize \hspace{4mm}\textbf{Ours M}}} &
\includegraphics[height=\turnheightnew]{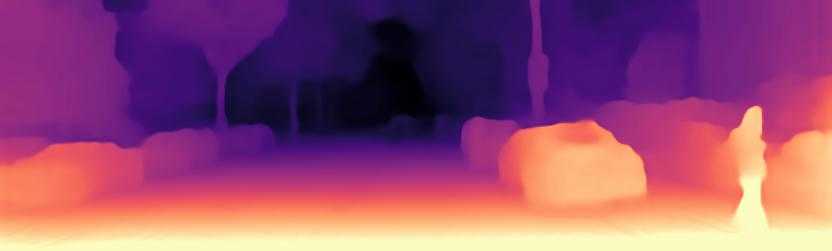} &
\includegraphics[height=\turnheightnew]{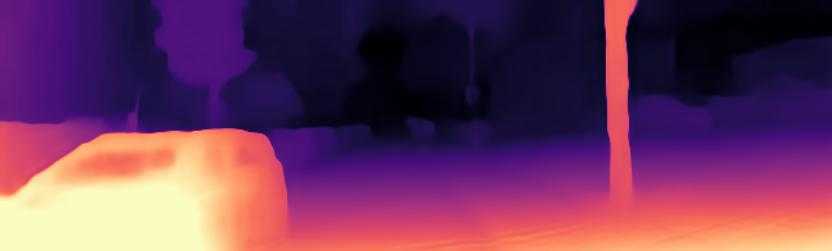} &
\includegraphics[height=\turnheightnew]{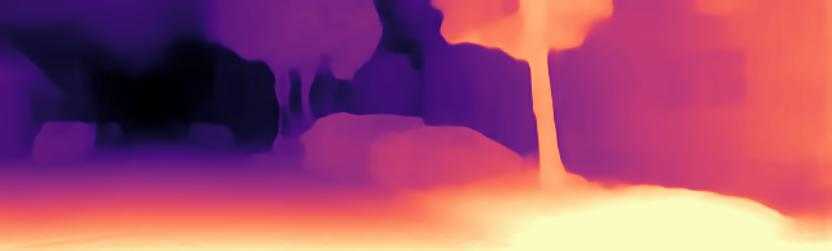} &
\includegraphics[height=\turnheightnew]{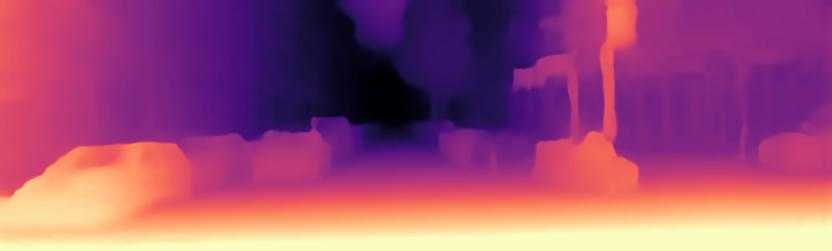} \\

\end{tabular}
    }
    \caption{{\bf Additional Qualitative Comparison.} A comparison of our method (\emph{last row}) with several other methods for monocular and stereo trained self supervised depth estimation.}
    \label{fig:supp_method_comparison}
\end{figure*}

\begin{figure*}[!htbp]
    \centering
    \resizebox{\textwidth}{!}{
    	\newcommand{\imlabel}[2]{\includegraphics[width=0.49\columnwidth]{#1}%
\raisebox{2pt}{\makebox[-2pt][r]{\footnotesize #2}}}
\newcommand{\turnheightnew}{0.195\columnwidth}
\centering

\begin{tabular}{@{\hskip 1mm}c@{\hskip 1mm}c@{\hskip 1mm}c}

\includegraphics[width=0.49\columnwidth]{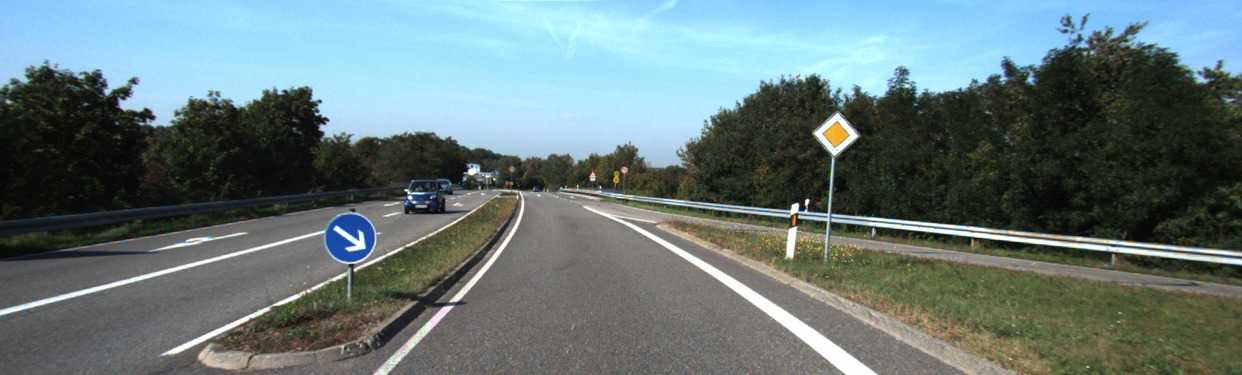} &
\includegraphics[width=0.49\columnwidth]{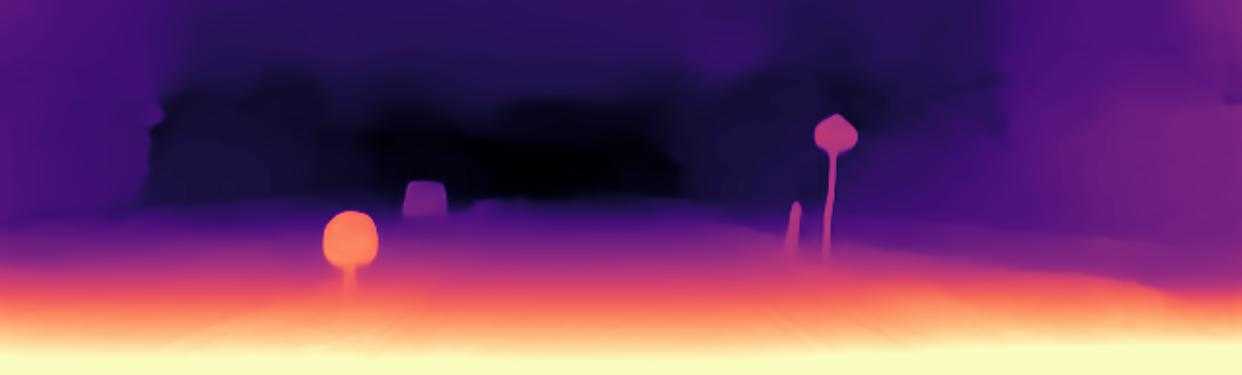} &
\includegraphics[width=0.49\columnwidth]{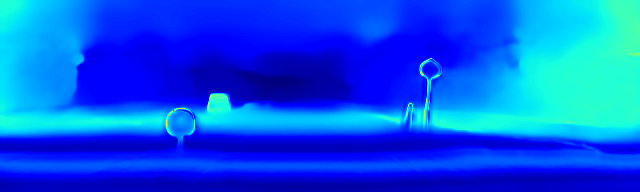} \\

\includegraphics[width=0.49\columnwidth]{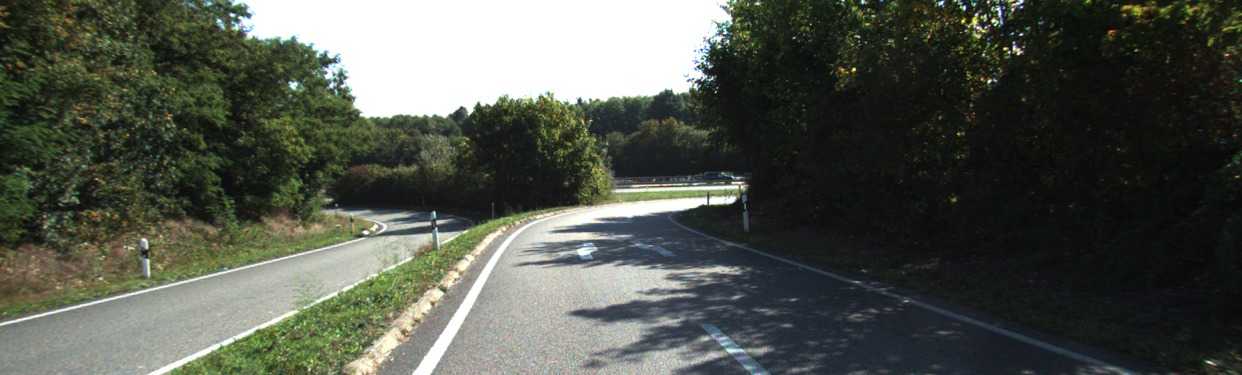} &
\includegraphics[width=0.49\columnwidth]{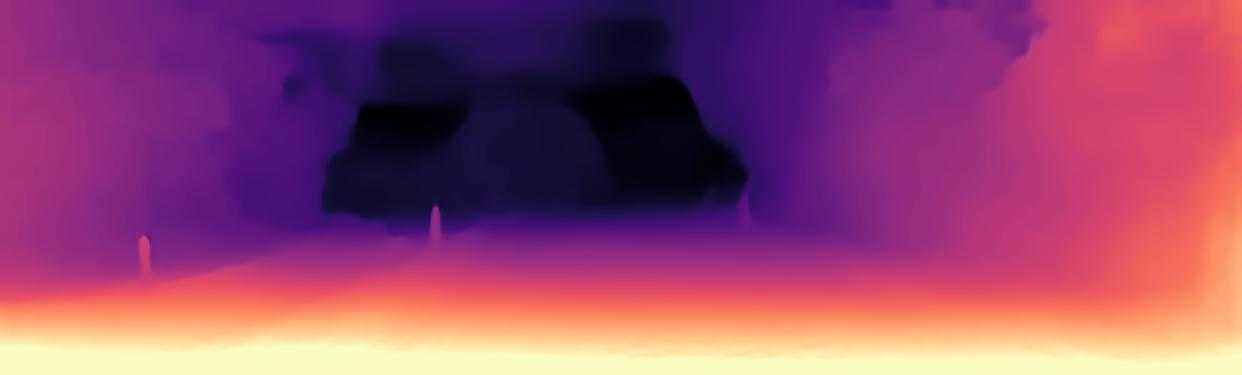} &
\includegraphics[width=0.49\columnwidth]{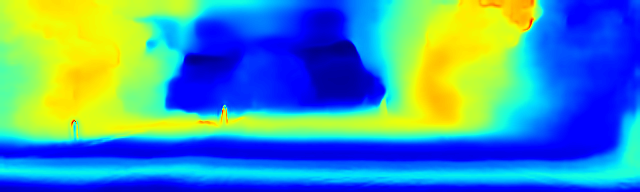} \\

\includegraphics[width=0.49\columnwidth]{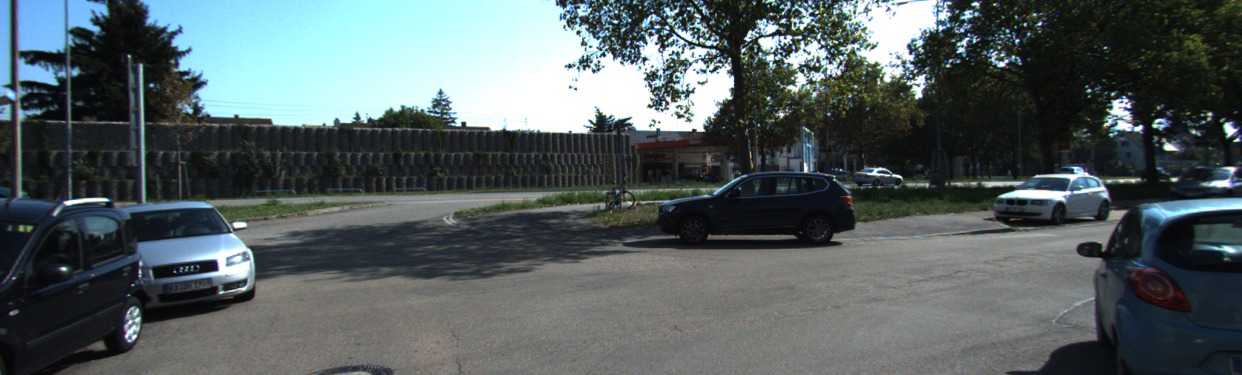} &
\includegraphics[width=0.49\columnwidth]{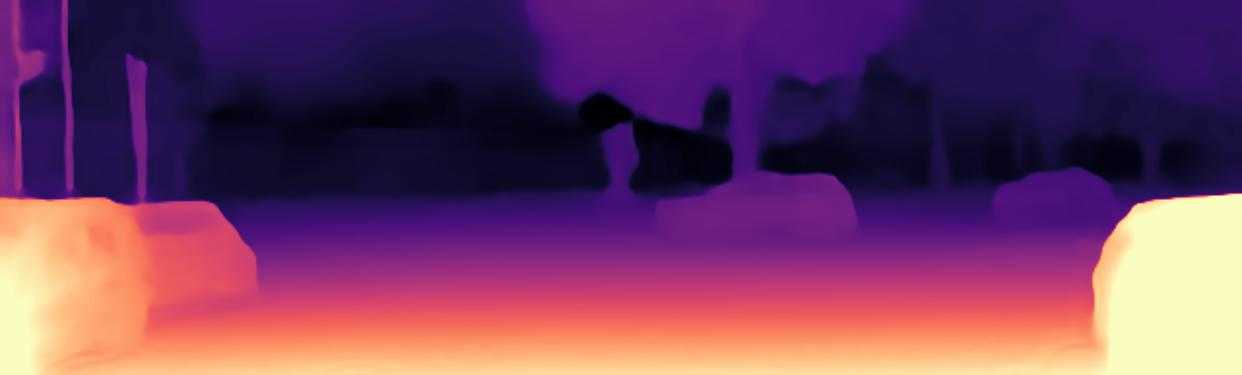} &
\includegraphics[width=0.49\columnwidth]{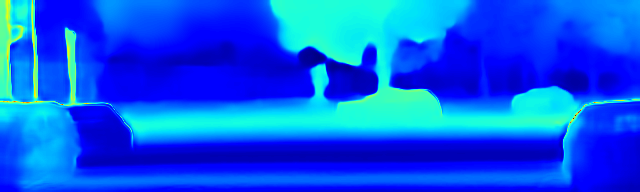} \\

\includegraphics[width=0.49\columnwidth]{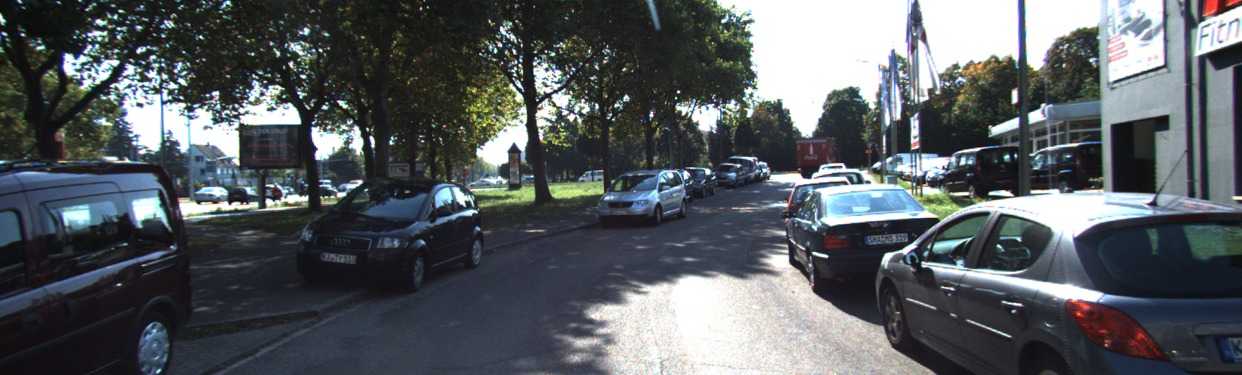} &
\includegraphics[width=0.49\columnwidth]{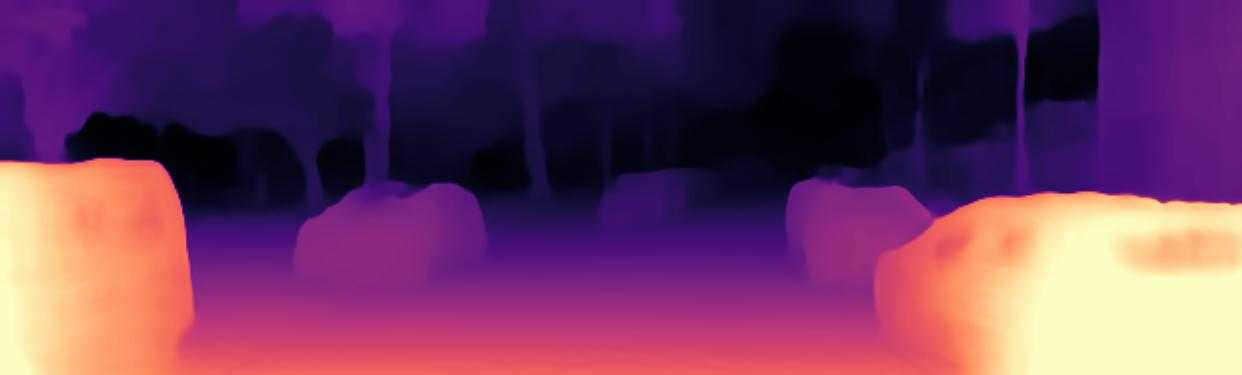} &
\includegraphics[width=0.49\columnwidth]{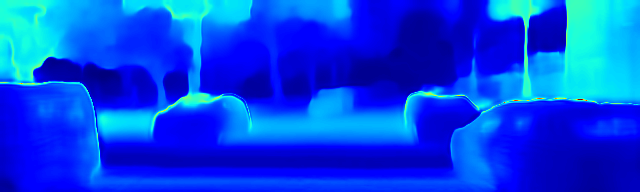} \\

\includegraphics[width=0.49\columnwidth]{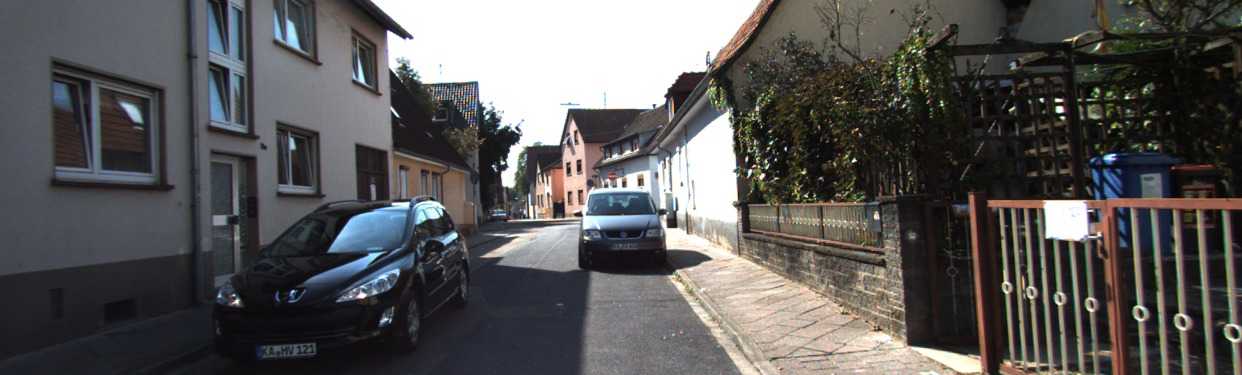} &
\includegraphics[width=0.49\columnwidth]{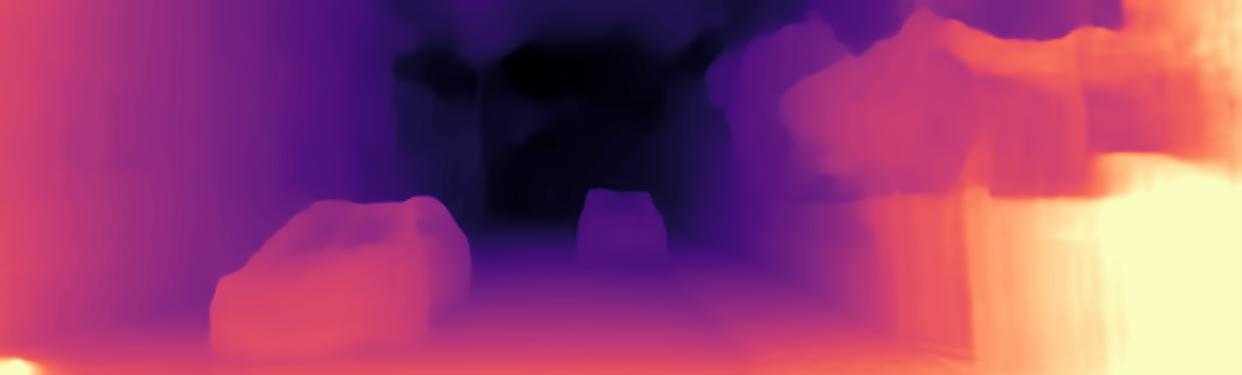} &
\includegraphics[width=0.49\columnwidth]{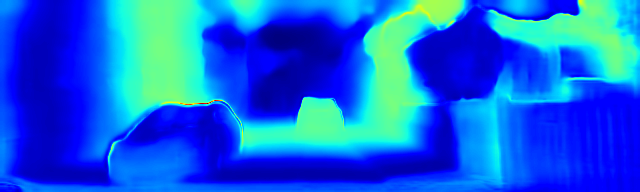}  \\

\includegraphics[width=0.49\columnwidth]{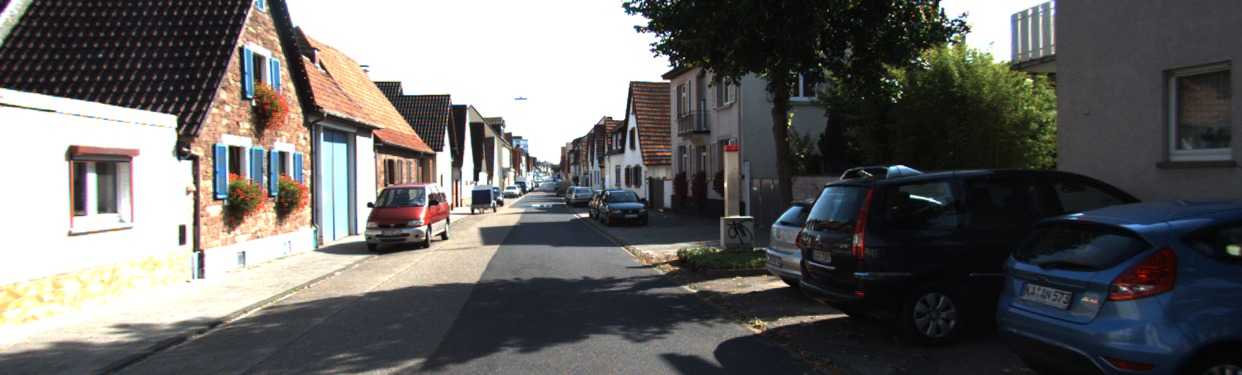} &
\includegraphics[width=0.49\columnwidth]{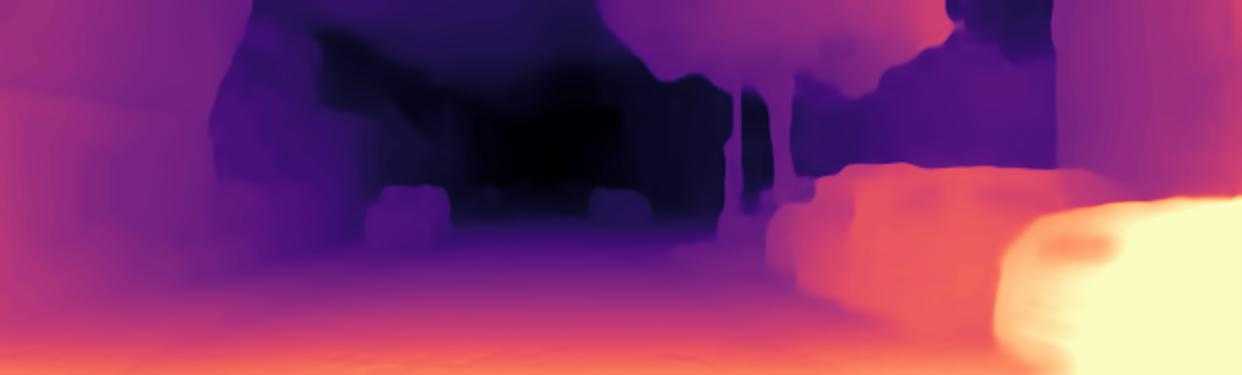} &
\includegraphics[width=0.49\columnwidth]{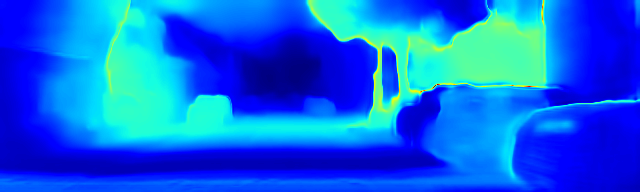}  \\

\includegraphics[width=0.49\columnwidth]{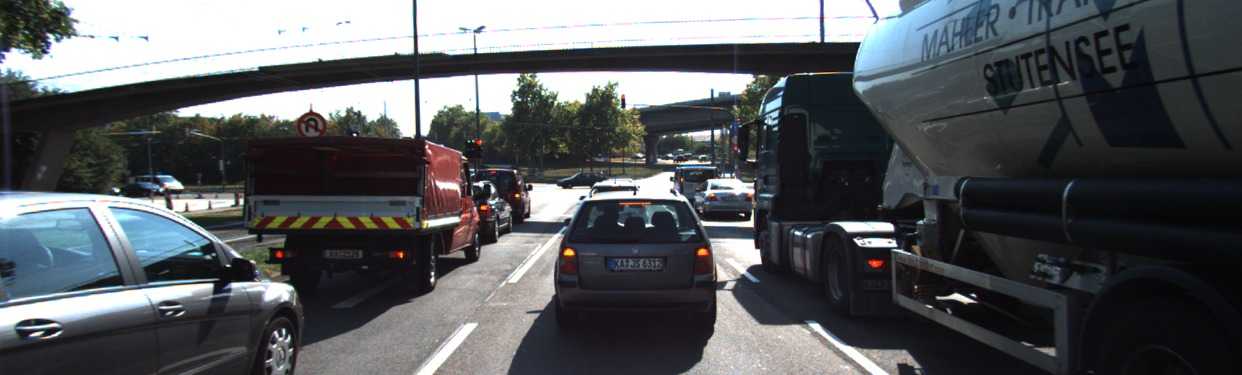} &
\includegraphics[width=0.49\columnwidth]{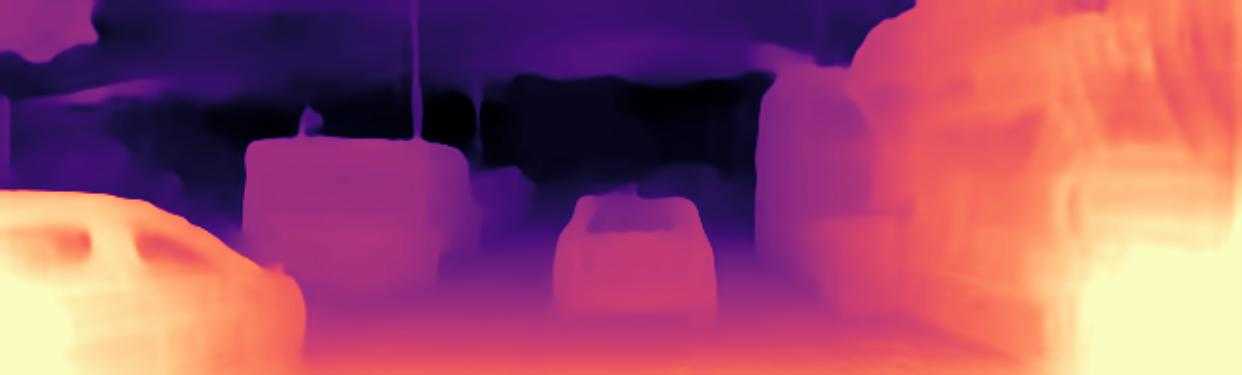} &
\includegraphics[width=0.49\columnwidth]{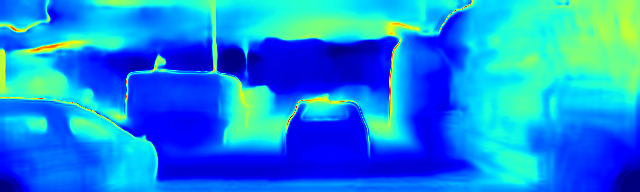}  \\

\end{tabular}
    }
    \caption{{\bf Additional uncertainty results} The Discrete Disparity Volume (DDV) allows us to compute pixel-wise depth uncertainty by measuring the variance across the disparity \emph{ray}. Left: Input Image, Middle: Depth prediction, Right: Uncertainty (Blue indicates areas of low uncertainty, green/red regions indicate areas of high/highest uncertainty).}
    \label{fig:supp_uncert}
\end{figure*}

\begin{figure*}[!htbp]
    \centering
    \resizebox{\textwidth}{!}{
    	\newcommand{\turnheightnew}{0.49\columnwidth}

\centering

\begin{tabular}{@{\hskip 2mm}c@{\hskip 2mm}c@{\hskip 2mm}c@{\hskip 2mm}c@{\hskip 2mm}c@{}}

\includegraphics[height=\turnheightnew]{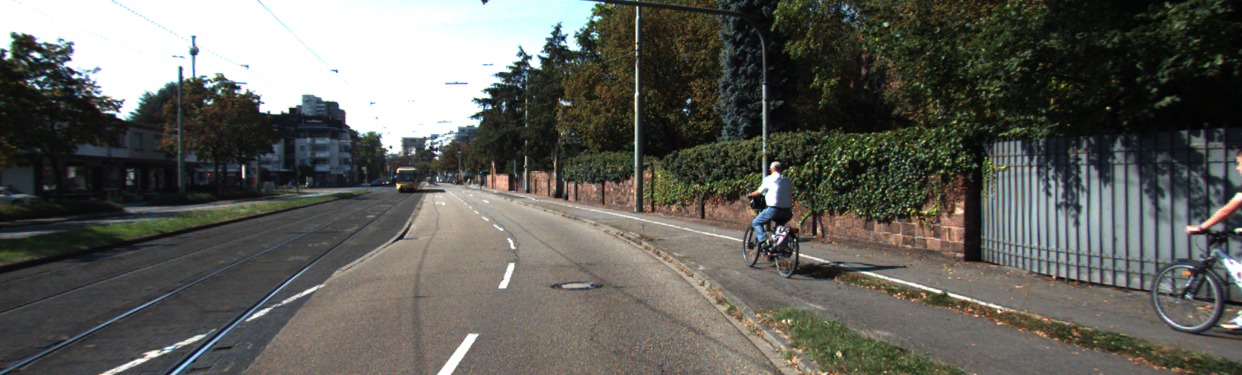} &
\includegraphics[height=\turnheightnew]{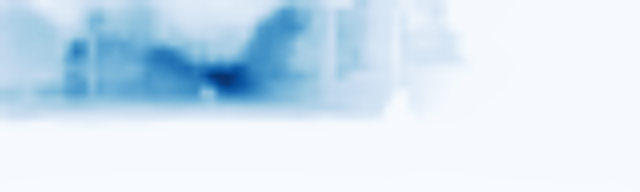} &
\includegraphics[height=\turnheightnew]{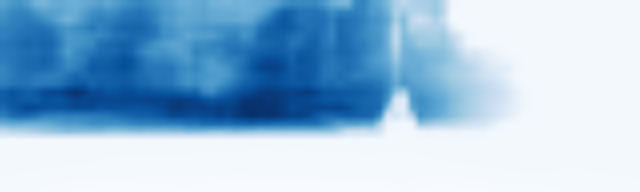} &
\includegraphics[height=\turnheightnew]{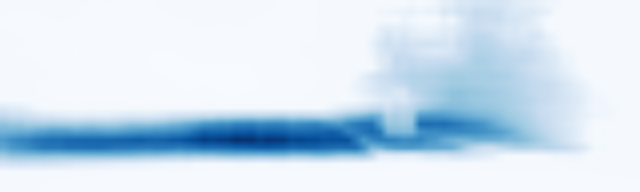} \\

\includegraphics[height=\turnheightnew]{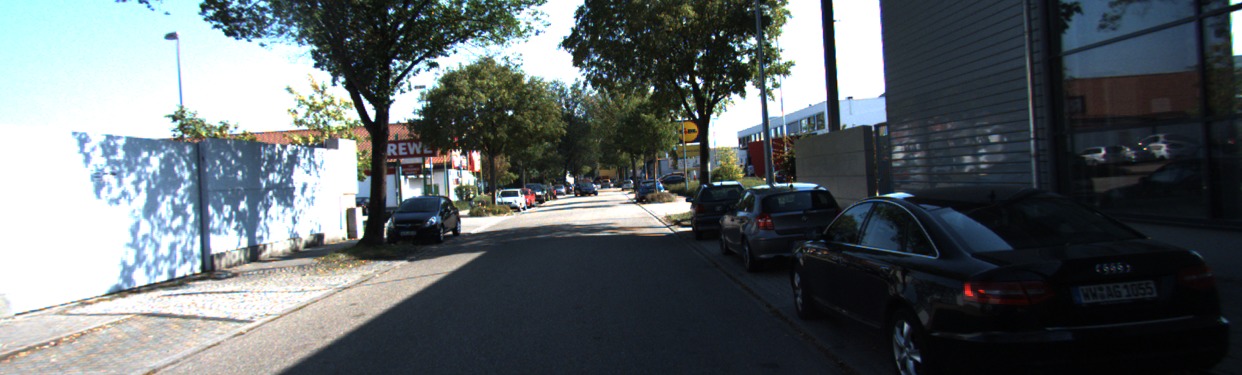} &
\includegraphics[height=\turnheightnew]{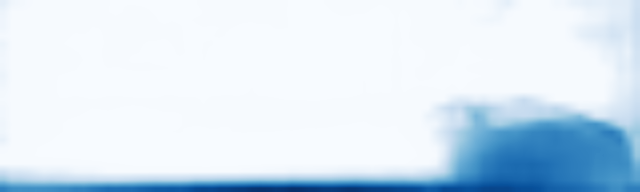} &
\includegraphics[height=\turnheightnew]{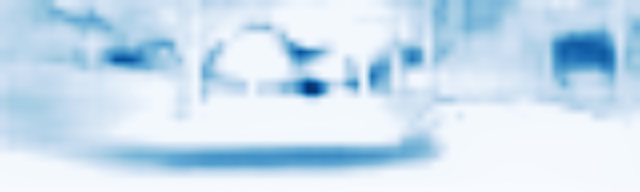} &
\includegraphics[height=\turnheightnew]{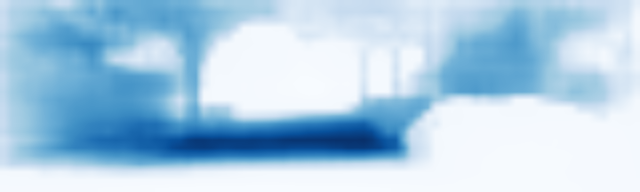} \\

\includegraphics[height=\turnheightnew]{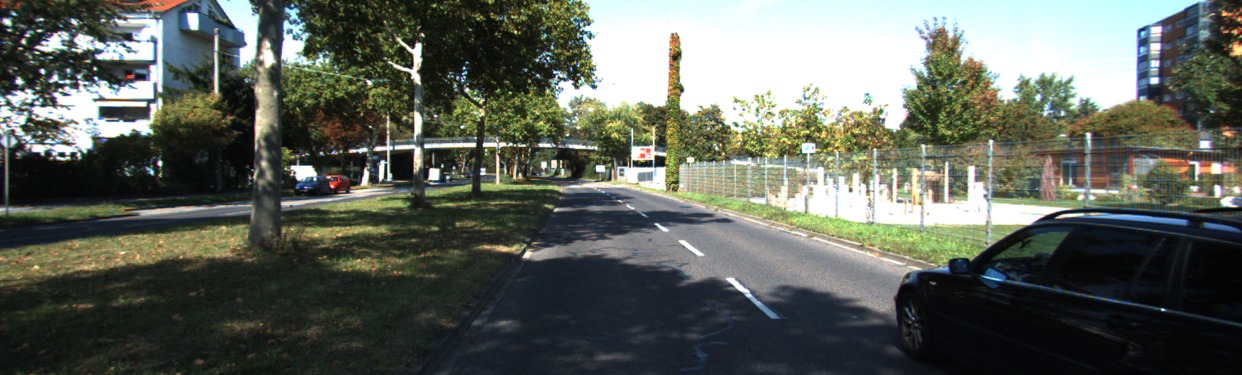} &
\includegraphics[height=\turnheightnew]{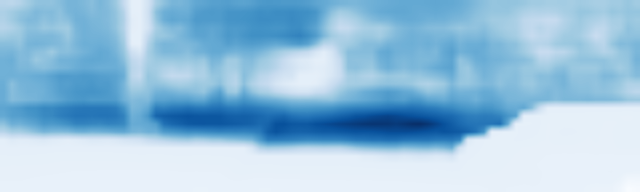} &
\includegraphics[height=\turnheightnew]{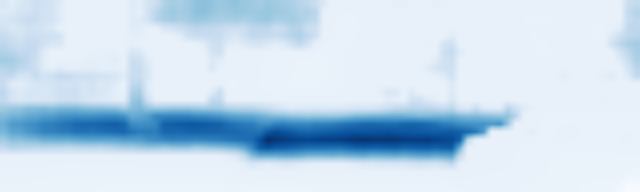} &
\includegraphics[height=\turnheightnew]{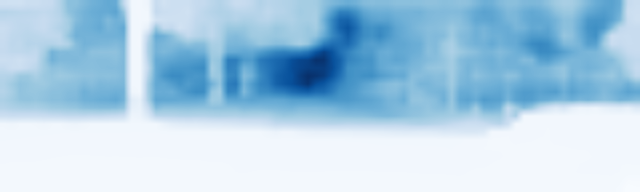} \\

\includegraphics[height=\turnheightnew]{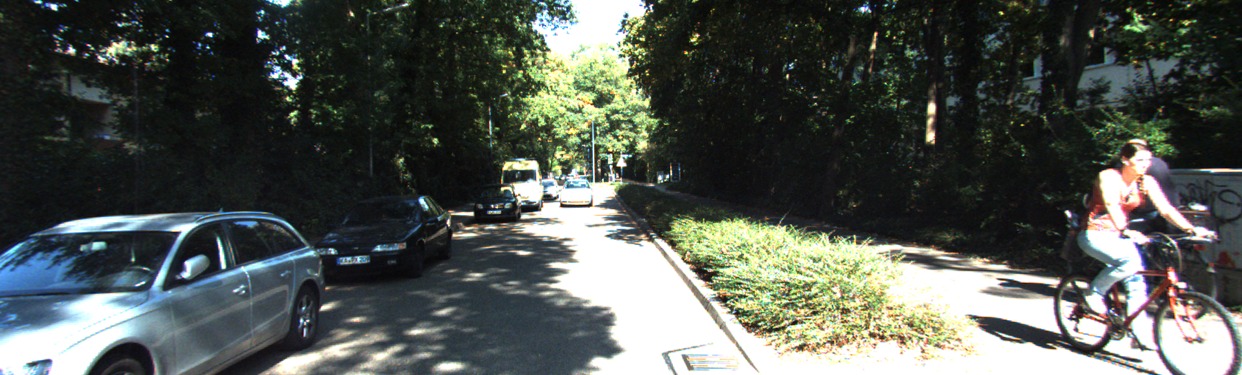} &
\includegraphics[height=\turnheightnew]{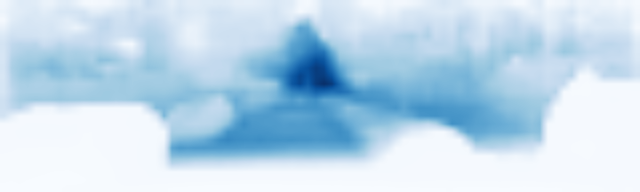} &
\includegraphics[height=\turnheightnew]{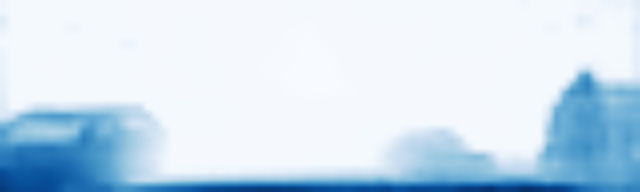} &
\includegraphics[height=\turnheightnew]{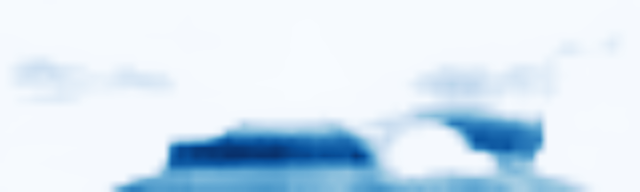} \\

\includegraphics[height=\turnheightnew]{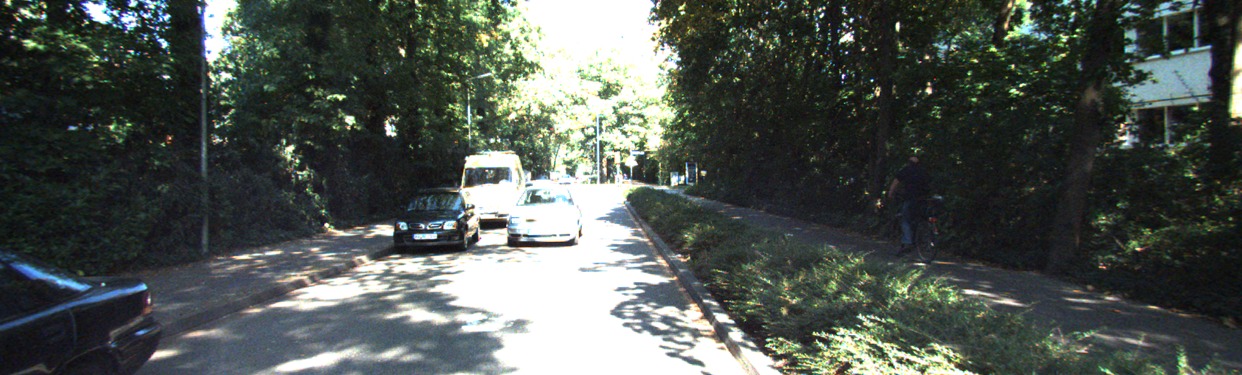} &
\includegraphics[height=\turnheightnew]{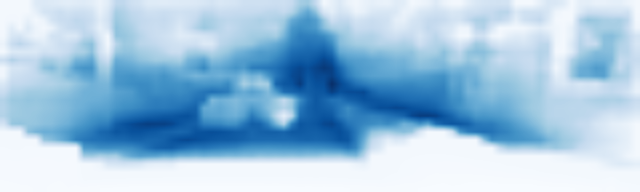} &
\includegraphics[height=\turnheightnew]{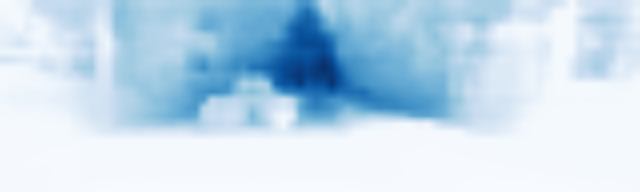} &
\includegraphics[height=\turnheightnew]{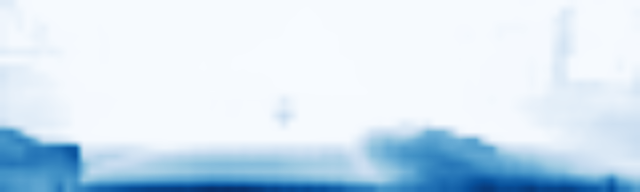} \\

\includegraphics[height=\turnheightnew]{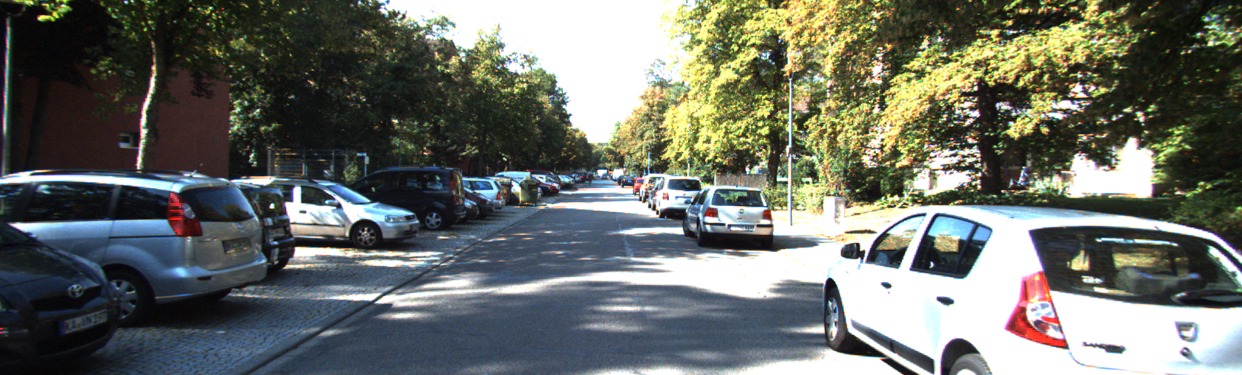} &
\includegraphics[height=\turnheightnew]{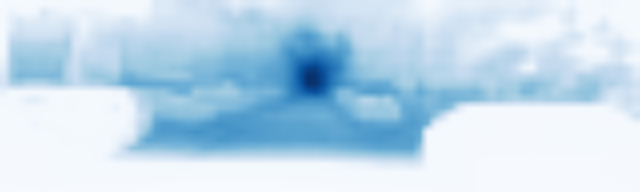} &
\includegraphics[height=\turnheightnew]{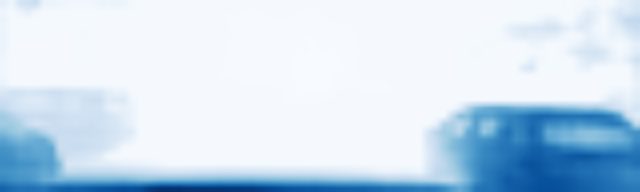} &
\includegraphics[height=\turnheightnew]{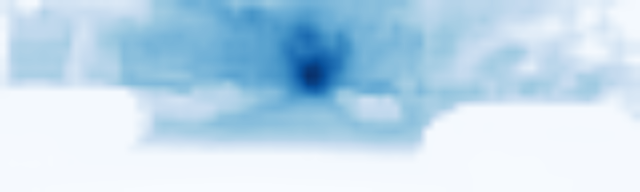} \\

\end{tabular}
    }
    \caption{{\bf Additional attention maps} selected at random from the output of context module (Blue indicates areas of high attention).}
    \label{fig:supp_attn}
\end{figure*}

In Figure \ref{fig:supp_method_comparison}, we present additional qualitative comparisons to multiple previous works. Our method produces sharper predictions for thin structures and complex shapes such as people.
In Figure \ref{fig:supp_uncert}, we show the uncertainty estimates for multiple images. As can been seen in the figure, areas of low contrast (row 2) correspond with areas of high uncertainty. Moreover, high uncertainty can also be observed in areas of unknown texture (row 7, right hand side). This area of the input image also demonstrates issues with texture copy artefacts~\cite{monodepth2} in the predicted depth. Additional attention maps are displayed in Figure \ref{fig:supp_attn}. The attention maps were selected at random from the 512 output channels in the context module.

\begin{table*}
  \centering
  
  \begin{tabular}{|l|c||c|c|c|c|c|c|c|}
  \hline
  Method & Train & \cellcolor{col1}Abs Rel & \cellcolor{col1}Sq Rel & \cellcolor{col1}RMSE  & \cellcolor{col1}RMSE log & \cellcolor{col2}$\delta < 1.25 $ & \cellcolor{col2}$\delta < 1.25^{2}$ & \cellcolor{col2}$\delta < 1.25^{3}$\\
  \hline 
Zhou \cite{zhou2017unsupervised}\textdagger & M & 0.176 & 1.532 & 6.129 & 0.244 & 0.758 & 0.921 & 0.971\\
Mahjourian \cite{mahjourian2018unsupervised} & M & 0.134 & 0.983 & 5.501 & 0.203 & 0.827 & 0.944 & 0.981\\
GeoNet \cite{geonet2018} & M  & 0.132 & 0.994 & 5.240 & 0.193 & 0.833 & 0.953 & 0.985\\
DDVO \cite{wang2017learning} & M  & 0.126 & 0.866 & 4.932 & 0.185 & 0.851 & 0.958 & 0.986\\ 
Ranjan \cite{ranjan2018adversarial}  & M & 0.123 & 0.881 & 4.834 & 0.181 & 0.860 & 0.959 & 0.985\\
EPC++ \cite{luo2018every} & M & 0.120 & 0.789 & 4.755 & 0.177 & 0.856 & 0.961 & 0.987\\
{Monodepth2} \cite{monodepth2} w/o pretraining & M &   {0.112} &   {0.715} &  {4.502} &  {0.167} &   \underline{0.876} &   {0.967} &   {0.990}  \\ %
{Monodepth2} \cite{monodepth2} & M &   \underline{ 0.090}  &   \underline{ 0.545}  &   \underline{ 3.942}  &   \underline{ 0.137}  &   \underline{ 0.914}  &   \underline{ 0.983}  &   \underline{0.995} \\ 

\rowcolor{gray!30} {\bf Ours} & M & {\bf 0.081} & {\bf 0.484}   & {\bf  3.716} & {\bf 0.126}    & {\bf  0.927}  & {\bf 0.985} & {\bf 0.996}   \\ 
\hline

Monodepth \cite{godard2017unsupervised} & S & 0.109 & 0.811 & 4.568 & 0.166 & 0.877 & 0.967 & 0.988\\
3net \cite{poggi20183net} (VGG) & S  &   0.119 &   0.920 &   4.824 &   0.182 &   0.856 &   0.957 &   0.985 \\ 
3net \cite{poggi20183net} (ResNet 50) & S  &   0.102 &   0.675 &   4.293 &   0.159 &   0.881 &   0.969 & \underline{0.991} \\ 
SuperDepth \cite{pillai2018superdepth} + pp & S & \underline{0.090} & \underline{0.542} & \underline{3.967} & \underline{0.144} & \underline{0.901} & \underline{0.976} & \textbf{0.993} \\

{Monodepth2} \cite{monodepth2} w/o pretraining& S &   0.110  &   0.849  &   4.580  &   0.173  &   0.875  &   0.962  &   0.986  \\
{Monodepth2} \cite{monodepth2}& S &   \textbf{0.085}  &   \textbf{0.537}  &   \textbf{3.868}  &   \textbf{0.139}  &   \textbf{0.912}  &   \textbf{0.979}  &   \textbf{0.993}  \\

\hline

Zhan FullNYU \cite{zhanst2018} & D*MS  & 0.130 & 1.520 & 5.184 & 0.205 & 0.859 & 0.955 & 0.981\\ 
EPC++ \cite{luo2018every} & MS & 0.123 & 0.754 & 4.453 & 0.172 & 0.863 & 0.964 & \underline{0.989}\\
{Monodepth2}\cite{monodepth2} w/o pretraining& MS & 
 \underline{0.107} & \underline{0.720} & \underline{4.345} & \underline{0.161} & \underline{0.890} & \underline{0.971} & \underline{0.989} \\
 
{Monodepth2}\cite{monodepth2}& MS & 
 \textbf{0.080} & \textbf{0.466} & \textbf{3.681} & \textbf{0.127} & \textbf{0.926} & \textbf{0.985} & \textbf{0.995} \\ 
 
 \hline

  \end{tabular}
  \caption{\textbf{Quantitative results on KITTI improved ground truth.} Comparison of existing methods to our own on the KITTI 2015 \cite{Geiger2012CVPR} using the improved ground truth \cite{uhrig2017sparse} of the Eigen test split \cite{eigen2015predicting}. The Best results are presented in {\bf bold} for each category, with second best results \underline{underlined}. The supervision level for each method is presented in the \emph{Train} column with; D -- Depth Supervision, D* -- Auxiliary depth supervision, S -- Self-supervised stereo supervision, M -- Self-supervised mono supervision. Results are presented without any post-processing \cite{godard2017unsupervised}, unless marked with -- + pp. If newer results are available on github, these are marked with -- \textdagger. Non-Standard resolutions are documented along with the method name.  Metrics indicated by red: \emph{lower is better}, Metrics indicated by blue: \emph{higher is better}}

  \label{tab:kitti_improved_gt}

\end{table*}

\end{appendices}

\end{document}